\documentclass[lettersize,journal]{IEEEtran}
\usepackage{amsmath}
\usepackage{times}
\usepackage{graphicx}
\usepackage[colorlinks=true, allcolors=blue]{hyperref}
\usepackage{booktabs}
\usepackage{array}
\usepackage{makecell}
\usepackage{pifont}
\usepackage{caption}
\usepackage{authblk}
\usepackage[T1]{fontenc}
\usepackage{multirow}
\usepackage{hyperref}
\usepackage{longtable}    

\usepackage{makecell}
\setcellgapes{4pt}
\makegapedcells

\usepackage[english]{babel}
\newtheorem{theorem}{Theorem}

\usepackage{array} 

\usepackage{amsmath}
\usepackage{amsfonts}  
\usepackage{tikz}
\usetikzlibrary{arrows.meta,positioning,calc}
\usepackage{graphicx}
\usepackage{pifont}
\usepackage[table]{xcolor}
\usepackage{multirow}
\usepackage{cite}
\usepackage{longtable} 

\usepackage[table]{xcolor}
\definecolor{mygreen}{rgb}{0.604, 0.98, 0.702}  
\definecolor{myred}{rgb}{1, 0.808, 0.808} 

\usepackage{comment}

\usepackage{array}
\usepackage{verbatim}
\usepackage{color}
\graphicspath{{eps/}}
\DeclareGraphicsExtensions{.eps}
\usepackage{ifpdf}
\ifpdf
\usepackage{epstopdf}
\fi
\usepackage{rotating}

\usepackage{cases}
\usepackage{diagbox}
\usepackage{lettrine}
\usepackage{textcomp}

\usepackage{amsthm}
\newtheorem{mytheorem}{Theorem}

\usepackage{tikz}
\newcommand{\circlednum}[1]{%
  \tikz[baseline=(char.base)]\node[shape=circle,draw,inner sep=0.4pt,fill=black,text=white,font=\bfseries\small] (char) {#1};%
}

\AtBeginDocument{%
  \fontdimen2\font=0.21em 
  \fontdimen3\font=0.1em  
  \fontdimen4\font=0.06em  
}

\begin{document}

\title{\textsc{OmniFuser}: Adaptive Multimodal Fusion for Service-Oriented Predictive Maintenance} 
\author{
        Ziqi~Wang,~\IEEEmembership{Student~Member,~IEEE,}
        Hailiang~Zhao,~\IEEEmembership{Member,~IEEE,}
        Yuhao Yang, Daojiang Hu, Cheng Bao,
        Mingyi Liu, Kai Di,
        Schahram~Dustdar,~\IEEEmembership{Fellow,~IEEE,}
        Zhongjie Wang,~\IEEEmembership{Member,~IEEE,}
        Shuiguang~Deng,~\IEEEmembership{Senior~Member,~IEEE}
\thanks{Hailiang Zhao is the corresponding author.}
\thanks{
    Ziqi Wang, Hailiang Zhao, Yuhao Yang, and Daojiang Hu are with the School of Software Technology, Zhejiang University. Emails: \{hliangzhao, wangziqi0312, yuhaoyang, daojianghu\}@zju.edu.cn.
}
\thanks{Cheng Bao is with the School of Computer Science and Technology, East China Normal University, Shanghai, China. Email: 18258681335@163.com.}
\thanks{Mingyi Liu and Zhongjie Wang are with the Faculty of Computing, Harbin Institute of Technology. Emails: \{liumy, rainy\}@hit.edu.cn. }
\thanks{Kai Di is with the Hangzhou School of Automation, Zhejiang Normal University. Email: dikai1994@zjnu.edu.cn.}
\thanks{Schahram Dustdar is with the Distributed Systems Group at the TU Wien and with ICREA at the UPF, Barcelona. Email: dustdar@dsg.tuwien.ac.at. }
\thanks{
    Shuiguang Deng is with the College of Computer Science and Technology, Zhejiang University. Email: dengsg@zju.edu.cn.
}
}



\maketitle

\begin{abstract}
Accurate and timely prediction of tool conditions is critical for intelligent manufacturing systems, where unplanned tool failures can lead to quality degradation and production downtime. In modern industrial environments, predictive maintenance is increasingly implemented as an intelligent service that integrates sensing, analysis, and decision support across production processes. To meet the demand for reliable and service-oriented operation, we present OmniFuser, a multimodal learning framework for predictive maintenance of milling tools that leverages both visual and sensor data. It performs parallel feature extraction from high-resolution tool images and cutting-force signals, capturing complementary spatiotemporal patterns across modalities. To effectively integrate heterogeneous features, OmniFuser employs a contamination-free cross-modal fusion mechanism that disentangles shared and modality-specific components, allowing for efficient cross-modal interaction. Furthermore, a recursive refinement pathway functions as an anchor mechanism, consistently retaining residual information to stabilize fusion dynamics. The learned representations can be encapsulated as reusable maintenance service modules, supporting both tool-state classification (e.g., Sharp, Used, Dulled) and multi-step force signal forecasting. Experiments on real-world milling datasets demonstrate that OmniFuser consistently outperforms state-of-the-art baselines, providing a dependable foundation for building intelligent industrial maintenance services. 
\end{abstract}



\section{Introduction}
\label{sec:introduction}
\IEEEPARstart{I}{n} modern manufacturing environments, ensuring the health and reliability of industrial equipment is critical to maintaining production efficiency, product quality, and economic competitiveness~\cite{1}. Among various components, milling tools play a pivotal role in precision machining, where even minor degradation can lead to dimensional inaccuracies, tool breakage, or unexpected downtime. Predicting the condition of milling tools in advance is essential to transitioning from reactive repairs to proactive interventions, thereby reducing maintenance costs and improving overall system resilience.
In recent years, predictive maintenance has evolved from a localized monitoring function into an \textit{intelligent industrial service}~\cite{TSC}, wherein sensing, analytics, and decision-making capabilities are encapsulated as reusable, interoperable service modules within smart manufacturing ecosystems. Such service-oriented predictive maintenance enables standardized, on-demand access to diagnostic and prognostic intelligence, allowing heterogeneous production units to dynamically discover, compose, and invoke maintenance services as part of integrated service workflows. However, the complex and dynamic nature of machining processes, which are characterized by noisy sensor streams, inconsistent wear progression, and variable operational conditions, poses significant challenges to the accuracy and robustness of such services \cite{liu1}. This motivates the need for \textit{intelligent, data-driven service models} that can holistically model equipment degradation and deliver actionable insights within industrial service frameworks.

Traditional approaches to tool condition monitoring predominantly rely on single-modality data, such as vibration, acoustic emission, or cutting-force signals \cite{r11,LSTM,gent2022}. These are typically processed using statistical features \cite{r11}, conventional machine learning models \cite{LSTM}, or spectral transforms \cite{gent2022} to classify wear states. While computationally efficient, such methods often fail to capture the full spectrum of degradation dynamics, especially under varying cutting loads and environmental interference~\cite{intr}. In parallel, vision-based techniques have been explored to detect surface wear or micro-cracks from tool images, leveraging handcrafted descriptors or deep convolutional networks \cite{textileCNN}. Yet, these approaches remain highly sensitive to lighting changes, occlusions, and viewpoint variations, limiting their reliability in real-world shop-floor settings~\cite{visiondraw}. Both signal-based and vision-based methods function as \textit{isolated} analytic services operating on disjoint data modalities, lacking mechanisms for cross-modal coordination or shared semantic understanding. To realize service-oriented predictive maintenance, it is imperative to integrate these heterogeneous perception services into a unified, collaborative intelligence layer. However, existing multimodal fusion strategies, which are often limited to early feature concatenation or late score averaging~\cite{multimodal}, fail to address fundamental challenges such as modality heterogeneity, asynchronous temporal dynamics, and cross-modal noise propagation. This underscores the need for an \textit{adaptive, contamination-aware fusion mechanism} that can align and synergize multimodal cues over time, serving as a foundational enabler for robust and reusable maintenance services.


To address these challenges, we propose \textsc{OmniFuser}, an omnidirectional multimodal fusion framework explicitly designed for \textit{service-oriented predictive maintenance}. \textsc{OmniFuser} performs parallel feature extraction from high-resolution tool images and time-series sensor signals, followed by a progressive fusion process that jointly aligns shared semantics while preserving modality-specific characteristics. Central to our design is the \textit{Contamination-free Cross-modal Fusion} ($\text{C}^2\text{F}$) mechanism, which disentangles shared and private representations to enable clean cross-modal interaction. A recursive refinement pathway further anchors the fusion dynamics by retaining residual information from original features, enhancing stability and robustness. The resulting multimodal representation is naturally encapsulable as a reusable maintenance service module, supporting dual prognostic tasks: (i) future tool-state classification (e.g., \textit{Sharp}, \textit{Used}, \textit{Dulled}) and (ii) multi-step forecasting of cutting-force signals. Experiments on real-world milling datasets demonstrate that \textsc{OmniFuser} consistently outperforms both unimodal baselines and state-of-the-art multimodal fusion methods, validating its efficacy as a core building block for intelligent industrial maintenance services. In summary, the main contributions of this work are as follows:
\begin{enumerate}
    \item We introduce \textsc{OmniFuser}, a novel service-oriented multimodal fusion framework for predictive maintenance, validated on milling tools as a representative industrial asset. The learned model can be directly encapsulated as a reusable, interoperable maintenance intelligence service, with potential applicability to other equipment monitored via heterogeneous data streams.
    \item We propose the $\text{C}^2\text{F}$ strategy, which enables progressive, bidirectional alignment of visual and sensor features while explicitly preserving modality-specific information. Coupled with a recursive refinement pathway that anchors fusion to original features, $\text{C}^2\text{F}$ mitigates information loss, temporal misalignment, and noise contamination.
    \item We conduct comprehensive experiments on two real-world datasets. \textsc{OmniFuser} achieves the best or second-best results at most horizons, yielding on average around 8-10\% lower MSE and MAE than recent baselines, and about 2\% higher accuracy in classification.
\end{enumerate}
The remainder of this paper is organized as follows. Section~\ref{RW} reviews related work. Section~\ref{Mo} presents the motivation, and the problem formulation is defined in Section~\ref{PF}. Section~\ref{PM} details the \textsc{OmniFuser} framework. Experimental evaluation is provided in Section~\ref{PE}, and Section~\ref{CF} concludes the paper.

\section{Related work} 
\label{RW}

\subsection{Predictive Maintenance with Single Modality}
Traditional predictive maintenance methods have predominantly relied on single-modality inputs to assess equipment health. In sensor-based approaches, early studies typically extract handcrafted statistical features or applied time-frequency transforms to characterize degradation signatures \cite{r11}. With the advent of deep learning, architectures such as convolutional neural networks (CNNs) and long short-term memory (LSTM) networks \cite{LSTM} have been employed to model temporal dependencies in sensor signals for remaining useful life estimation. Gent \textit{et al.}~\cite{gent2022} demonstrate a wireless instrumented tool holder capable of acquiring high-frequency acceleration data near the workpiece interface. By combining spectral analysis with a degradation index, their system successfully detects both progressive tool wear and abrupt cutting edge breakouts during an industrial trial. From a service-oriented perspective, these sensor-based models represent early attempts to encapsulate monitoring and diagnostic capabilities as independent analytical components within intelligent maintenance services. However, such techniques often struggle to characterize the complex degradation patterns that arise from varying machining parameters. Moreover, their performance can be highly sensitive to background noise and signal perturbations, resulting in limited robustness and poor generalization to unseen conditions. This poses challenges to reliable service delivery in dynamic industrial environments.

In parallel, image-based methods have been explored for detecting tool wear by analyzing images of the tool surface. Traditional approaches rely on handcrafted descriptors (\textit{e.g.}, texture, edge features), whereas more recent techniques employ deep CNNs to learn wear patterns. For instance, Muruganandham \textit{et al.}~\cite{textileCNN} develop an industrial defect detection framework in the textile domain, in which high-resolution images are captured under controlled lighting and processed through CNN architectures to extract multiscale texture and structural features. The trained network can classify fine-grained defect categories, achieving robust detection despite variations in defect size, shape, and color. A similar paradigm can be adapted to tool condition monitoring by capturing images of cutting edges to identify subtle wear signatures and delivering visual diagnostics as part of intelligent maintenance services. However, image-based methods are heavily influenced by external factors such as lighting variability, and static image analysis fails to capture dynamic changes in the cutting process that are critical for early failure prediction.

While single-modality methods can provide valuable insights into equipment health, their scope is inherently constrained by the information available within a single data source. The absence of cross-domain complementary cues limits their ability to capture the multifaceted nature of degradation processes. These shortcomings have driven growing interest in multimodal fusion strategies, which seek to integrate heterogeneous data streams to improve predictive accuracy and enhance robustness under variable operating conditions. This serves as a key enabler for intelligent and service-oriented predictive maintenance frameworks.

\subsection{Multimodal Fusion Methods for Equipment Monitoring}
Multimodal approaches integrate heterogeneous data sources to leverage their complementary characteristics. Early fusion strategies typically combine raw or low-level features through concatenation before feeding them into a joint network. Zeng \textit{et al.}~\cite{Zeng2006} develop a multimodal sensing framework for high-speed milling that integrates force, vibration, and acoustic emission sensors. Specifically, synchronized measurements from the three sensors are first preprocessed and transformed via wavelet decomposition to obtain time-frequency representations. The resulting spectral energy distributions from all modalities are then concatenated into a unified feature vector, enabling the identification of frequency bands closely associated with flank wear progression. However, early fusion methods often overlook the inherent differences in spatiotemporal structures across modalities, making them susceptible to feature redundancy and conflicts. Late fusion methods adopt a more modular strategy, where each modality is processed independently, and the final decisions are merged using score averaging or majority voting. These methods lack deep cross-modal interaction and fail to learn shared representations effectively, often limiting their predictive performance.

To address these challenges, recent studies have introduced attention-based fusion mechanisms to improve cross-modal interaction. For example, Low-rank Cross-modal Interaction Fusion \cite{Wang} performs deep cross-attention and low-rank interaction between the modality-specific feature sets, enabling effective exploitation of complementary information while mitigating redundancy. Guan \textit{et al.}~\cite{guan} introduce a Transformer-based multimodal framework, combining visual inputs with sensor and motion information. It uses a two-phase optimization strategy to strengthen the temporal association between observed and future video segments. However, these methods do not explicitly distinguish modality-specific representations and instead project all features into a shared embedding space for fusion, which may lead to temporal misalignment and loss of modality-specific information.


Table \ref{tab:related_work} shows the contrastive analysis of different studies. Different from the above works, \textsc{OmniFuser} presents a comprehensive and service-oriented framework for stepwise predictive maintenance of milling tools. It incorporates a three-stage $\text{C}^2\text{F}$ strategy that achieves contamination-free and low-complexity fusion by incrementally aligning and integrating multimodal features. $\text{C}^2\text{F}$ is distinguished by three aspects of novelty: (1) It defines contamination as cross-modal redundancy and enforces orthogonality between shared and private subspaces to preserve modality-specific cues; (2) an efficient proxy-based cross-modal attention mechanism that approximates the dominant interaction subspace with adaptive landmarks, achieving fidelity at reduced complexity; and (3) a recursive refinement strategy that anchors each fusion iteration to the original modality features, stabilizing representation updates and preventing information drift. These ensure consistent performance when deployed as an intelligent maintenance service module within industrial systems.

\begin{table*}[htbp!]
\centering
\caption{Contrastive Analysis of Different Studies (\ding{51}: involved; \ding{55}: not involved)}
\label{tab:related_work}
\resizebox{\textwidth}{!}{
\begin{tabular}{@{}lcccccccccccccc@{}}
\toprule
\multirow{2}{*}{\textbf{Related work}} 
& \multicolumn{3}{c}{\textbf{Modality type}} 
& \multicolumn{2}{c}{\textbf{Feature design}} 
& \multicolumn{4}{c}{\textbf{Fusion strategy}} 
& \multicolumn{3}{c}{\textbf{Cross-modal alignment}} 
& \multicolumn{2}{c}{\textbf{Application dimension}} \\
\cmidrule(lr){2-4} \cmidrule(lr){5-6} \cmidrule(lr){7-10} \cmidrule(lr){11-13} \cmidrule(lr){14-15}
 & Sensor-only & Image-only & Multimodal 
 & Handcrafted & Deep learning 
 & Early & Late & Attention-based & Progressive 
 & Noise robustness & Feature preservation & Generalization 
 & Tool wear detection & RUL estimation \\
\midrule
Alexandrina \textit{et al.} \cite{LSTM} 
& \cellcolor{mygreen}\ding{51} & \cellcolor{myred}\ding{55} & \cellcolor{myred}\ding{55} & \cellcolor{myred}\ding{55} & \cellcolor{mygreen}\ding{51} & \cellcolor{myred}\ding{55} & \cellcolor{myred}\ding{55} & \cellcolor{myred}\ding{55} & \cellcolor{myred}\ding{55} 
& \cellcolor{mygreen}\ding{51} & \cellcolor{myred}\ding{55} & \cellcolor{myred}\ding{55} 
& \cellcolor{myred}\ding{55} & \cellcolor{mygreen}\ding{51} \\

Gent \textit{et al.} \cite{gent2022} 
& \cellcolor{mygreen}\ding{51} & \cellcolor{myred}\ding{55} & \cellcolor{myred}\ding{55} 
& \cellcolor{myred}\ding{55} & \cellcolor{mygreen}\ding{51} 
& \cellcolor{myred}\ding{55} & \cellcolor{myred}\ding{55} & \cellcolor{myred}\ding{55} & \cellcolor{myred}\ding{55} 
& \cellcolor{myred}\ding{55} & \cellcolor{myred}\ding{55} & \cellcolor{myred}\ding{55} 
& \cellcolor{mygreen}\ding{51} & \cellcolor{mygreen}\ding{51} \\

Muruganandham \textit{et al.} \cite{textileCNN} 
& \cellcolor{myred}\ding{55} & \cellcolor{mygreen}\ding{51} & \cellcolor{myred}\ding{55} 
& \cellcolor{myred}\ding{55} & \cellcolor{mygreen}\ding{51} 
& \cellcolor{myred}\ding{55} & \cellcolor{myred}\ding{55} & \cellcolor{myred}\ding{55} & \cellcolor{myred}\ding{55} 
& \cellcolor{myred}\ding{55} & \cellcolor{myred}\ding{55} & \cellcolor{myred}\ding{55} 
& \cellcolor{mygreen}\ding{51} & \cellcolor{myred}\ding{55} \\

Zeng \textit{et al.} \cite{Zeng2006} 
& \cellcolor{myred}\ding{55} & \cellcolor{myred}\ding{55} & \cellcolor{mygreen}\ding{51} 
& \cellcolor{mygreen}\ding{51} & \cellcolor{myred}\ding{55} 
& \cellcolor{mygreen}\ding{51} & \cellcolor{myred}\ding{55} & \cellcolor{myred}\ding{55} & \cellcolor{myred}\ding{55} 
& \cellcolor{myred}\ding{55} & \cellcolor{myred}\ding{55} & \cellcolor{mygreen}\ding{51} 
& \cellcolor{myred}\ding{55} & \cellcolor{mygreen}\ding{51} \\

Bi \textit{et al.} \cite{Wang} 
& \cellcolor{myred}\ding{55} & \cellcolor{myred}\ding{55} & \cellcolor{mygreen}\ding{51} 
& \cellcolor{myred}\ding{55} & \cellcolor{mygreen}\ding{51} 
& \cellcolor{myred}\ding{55} & \cellcolor{myred}\ding{55} & \cellcolor{mygreen}\ding{51} & \cellcolor{myred}\ding{55} 
& \cellcolor{myred}\ding{55} & \cellcolor{mygreen}\ding{51} & \cellcolor{myred}\ding{55} 
& \cellcolor{myred}\ding{55} & \cellcolor{mygreen}\ding{51} \\

Guan \textit{et al.} \cite{guan} 
& \cellcolor{myred}\ding{55} & \cellcolor{myred}\ding{55} & \cellcolor{mygreen}\ding{51} 
& \cellcolor{myred}\ding{55} & \cellcolor{mygreen}\ding{51} 
& \cellcolor{myred}\ding{55} & \cellcolor{myred}\ding{55} & \cellcolor{mygreen}\ding{51} & \cellcolor{mygreen}\ding{51} 
& \cellcolor{mygreen}\ding{51} & \cellcolor{myred}\ding{55}  & \cellcolor{myred}\ding{55} 
& \cellcolor{myred}\ding{55} & \cellcolor{mygreen}\ding{51} \\

Truchan \textit{et al.} \cite{late} 
& \cellcolor{myred}\ding{55} & \cellcolor{myred}\ding{55} & \cellcolor{mygreen}\ding{51} 
& \cellcolor{myred}\ding{55} & \cellcolor{mygreen}\ding{51} 
& \cellcolor{myred}\ding{55} & \cellcolor{mygreen}\ding{51} & \cellcolor{myred}\ding{55} & \cellcolor{myred}\ding{55} 
& \cellcolor{myred}\ding{55} & \cellcolor{mygreen}\ding{51} & \cellcolor{mygreen}\ding{51} 
& \cellcolor{myred}\ding{55} & \cellcolor{mygreen}\ding{51} \\

\textbf{Our work}   
& \cellcolor{myred}\ding{55} & \cellcolor{myred}\ding{55} & \cellcolor{mygreen}\ding{51} 
& \cellcolor{myred}\ding{55} & \cellcolor{mygreen}\ding{51} 
& \cellcolor{mygreen}\ding{51} & \cellcolor{myred}\ding{55} & \cellcolor{mygreen}\ding{51} & \cellcolor{mygreen}\ding{51} 
& \cellcolor{mygreen}\ding{51} & \cellcolor{mygreen}\ding{51} & \cellcolor{mygreen}\ding{51} 
& \cellcolor{mygreen}\ding{51} & \cellcolor{mygreen}\ding{51} \\
\bottomrule
\end{tabular}}
\end{table*}

\section{Motivation}
\label{Mo}



\begin{figure}[!htb]
\centering
\includegraphics[width=1\linewidth]{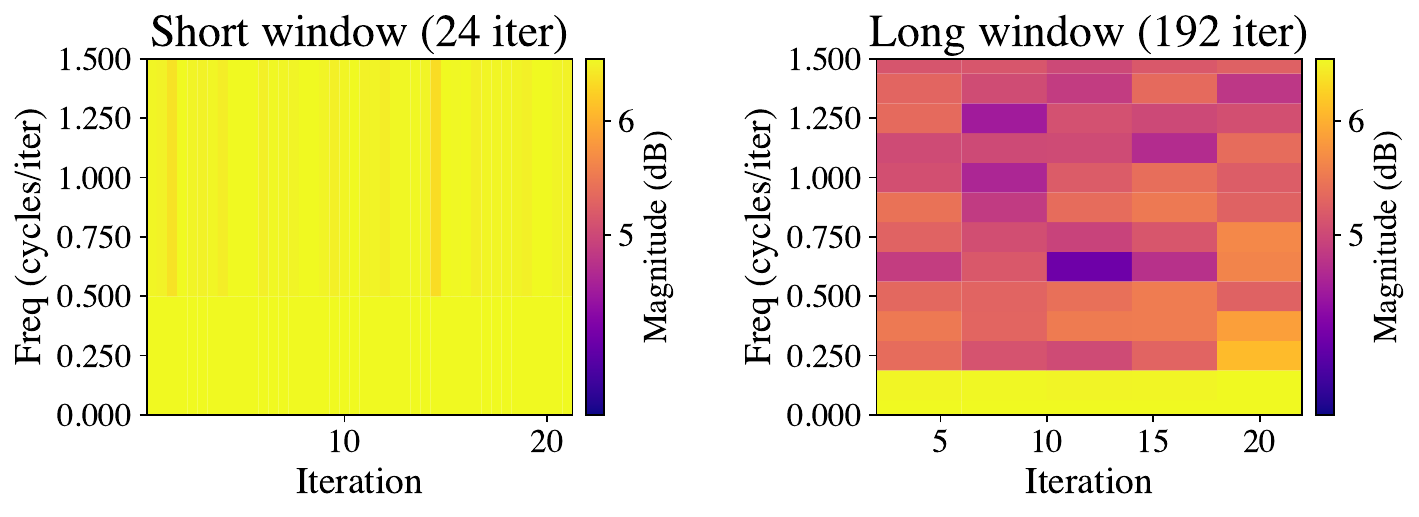}
\caption{Time-frequency analysis of cutting force signals.}
\label{mo1}
\end{figure}

In service-oriented predictive maintenance, the reliability of an intelligent maintenance service hinges on its ability to accurately capture both the \textit{slow degradation trends} and \textit{rapid operational dynamics} of industrial equipment. However, as illustrated in Fig.~\ref{mo1}, cutting-force signals from milling processes exhibit \textit{cross-scale temporal structures}: long-window time-frequency analysis reveals two stable low-frequency components that correspond to gradual tool wear, whereas short-window analysis, while offering better temporal localization, fails to resolve these slow variations due to limited frequency resolution. This duality implies that single-resolution models either oversmooth critical degradation cues or miss long-term evolution patterns, leading to unreliable service predictions.

\begin{figure}[!htb]
\centering
\includegraphics[width=1\linewidth]{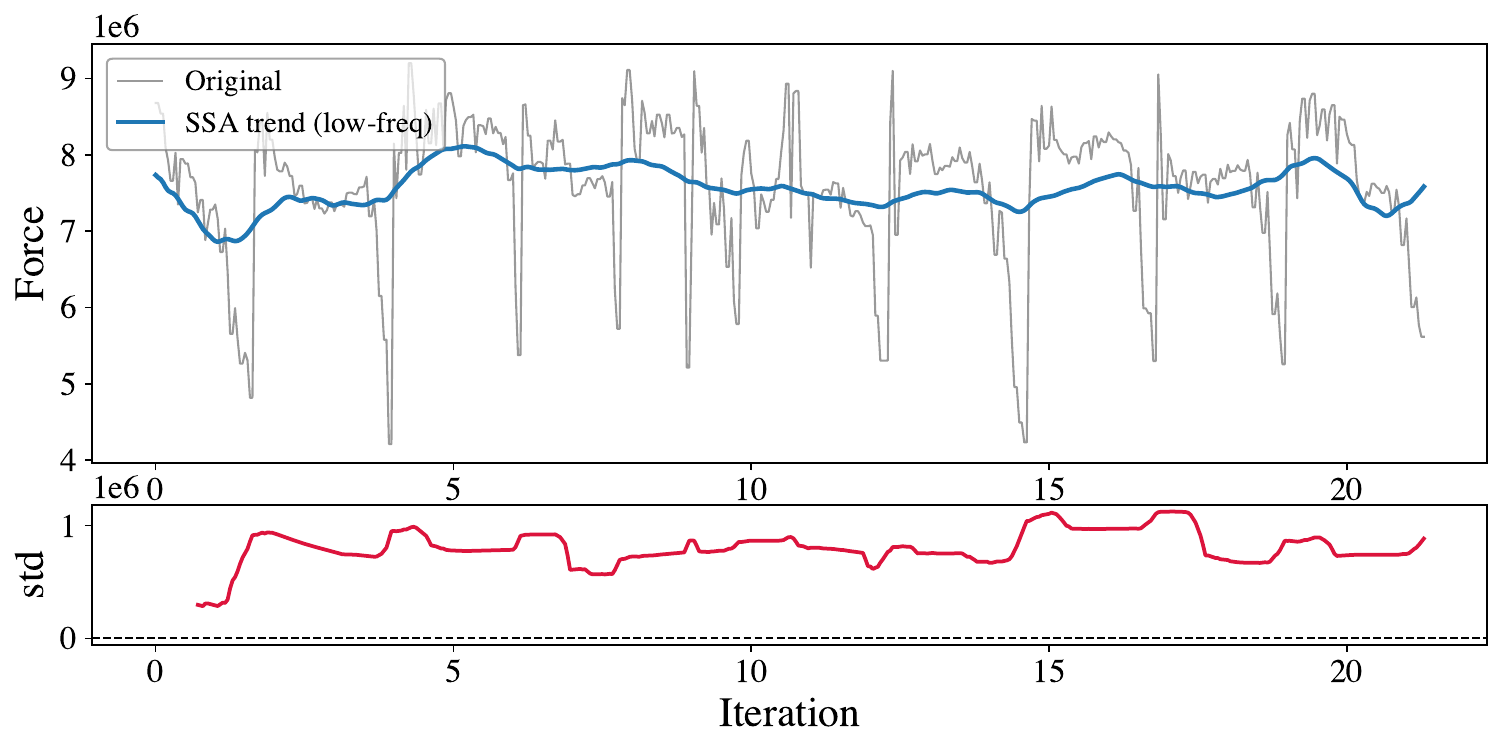}
\caption{Decomposition into trend and residual components.}
\label{mo2}
\end{figure}

\begin{figure}[!htb]
\centering
\includegraphics[width=1\linewidth]{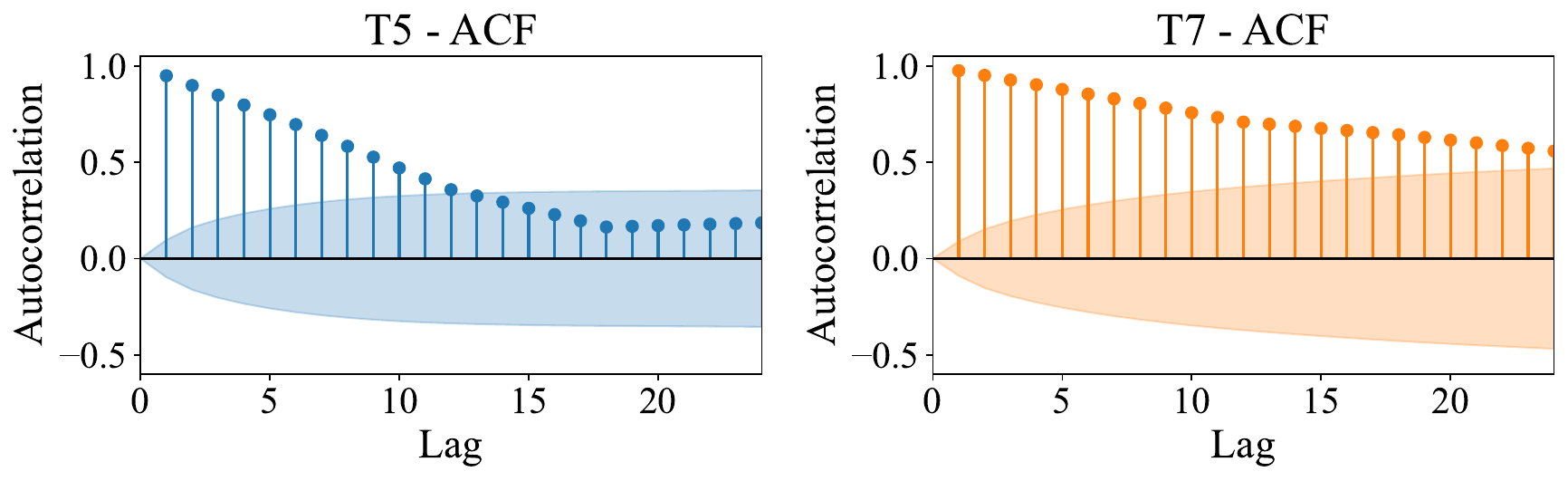}
\caption{Autocorrelation analysis on two milling tools.}
\label{mo3}
\end{figure}

To quantify this structure, we apply singular spectrum analysis (SSA) to decompose the force signal into trend and residual components. As shown in Fig.~\ref{mo2}, only the top 4 low-rank SSA components (explaining 18\% of the total variance) suffice to reconstruct the global wear trend, while the remaining 82\% of signal energy resides in high-frequency residuals. This energy distribution strongly supports a \textit{low-dimensional trend subspace} hypothesis: the essential degradation trajectory is compact, whereas transient disturbances, cutting vibrations, and noise dominate the high-frequency band. Consequently, a robust maintenance service must explicitly separate and model these two regimes to avoid conflating slow wear with short-term fluctuations. Furthermore, autocorrelation analysis in Fig.~\ref{mo3} shows that force signals from both tools exhibit rapid decay beyond a lag of approximately 17 steps, indicating limited long-range temporal memory. This confirms that sensor data alone primarily captures \textit{transient dynamics} but lacks persistent markers of cumulative wear. In contrast, visual observations of the tool edge, acquired after each cutting pass, provide direct, interpretable evidence of surface degradation (e.g., flank wear, chipping) that evolves slowly and is largely invariant within a single machining cycle. 

These observations motivate modality complementarity: the sensor stream offers high-frequency operational context, while the image stream anchors the service's understanding of physical wear state. Ignoring either modality risks incomplete or biased prognostics. Therefore, an effective maintenance service must integrate these asynchronous, heterogeneous data streams through a fusion mechanism that respects their distinct temporal semantics and noise characteristics.

\section{Problem Formulation}
\label{PF}
We formulate predictive maintenance as an intelligent service that delivers dual prognostic capabilities through a standardized multimodal interface. Specifically, the service receives a time-aligned observation window of length $T$ consisting of: (i) a temporal sensor sequence $\mathbf{X} = [\mathbf{x}_1, \dots, \mathbf{x}_T] \in \mathbb{R}^{T \times C}$, where $\mathbf{x}_t \in \mathbb{R}^C$ is the sensor reading at time $t$; and (ii) a visual sequence $\mathbf{I} = [\mathbf{I}_1, \dots, \mathbf{I}_T]$, where $\mathbf{I}_t \in \mathbb{R}^{H \times W}$ is the tool surface image at time $t$. In response, the service outputs a $P$-step-ahead prognosis: (i) predicted force signals $\hat{\mathbf{Y}} = [\hat{\mathbf{y}}_{T+1}, \dots, \hat{\mathbf{y}}_{T+P}] \in \mathbb{R}^{P \times C}$ for process monitoring, and (ii) discrete wear states $\hat{\mathbf{S}} = [\hat{s}_{T+1}, \dots, \hat{s}_{T+P}]$ with $\hat{s}_t \in \{\text{Sharp}, \text{Used}, \text{Dulled}\}$ for maintenance decision support. 

\section{The \textsc{OmniFuser} Framework}
\label{PM}
\subsection{Overall Architecture}
\begin{figure*}    
[!ht]
\centering
\includegraphics[width=0.97\linewidth]{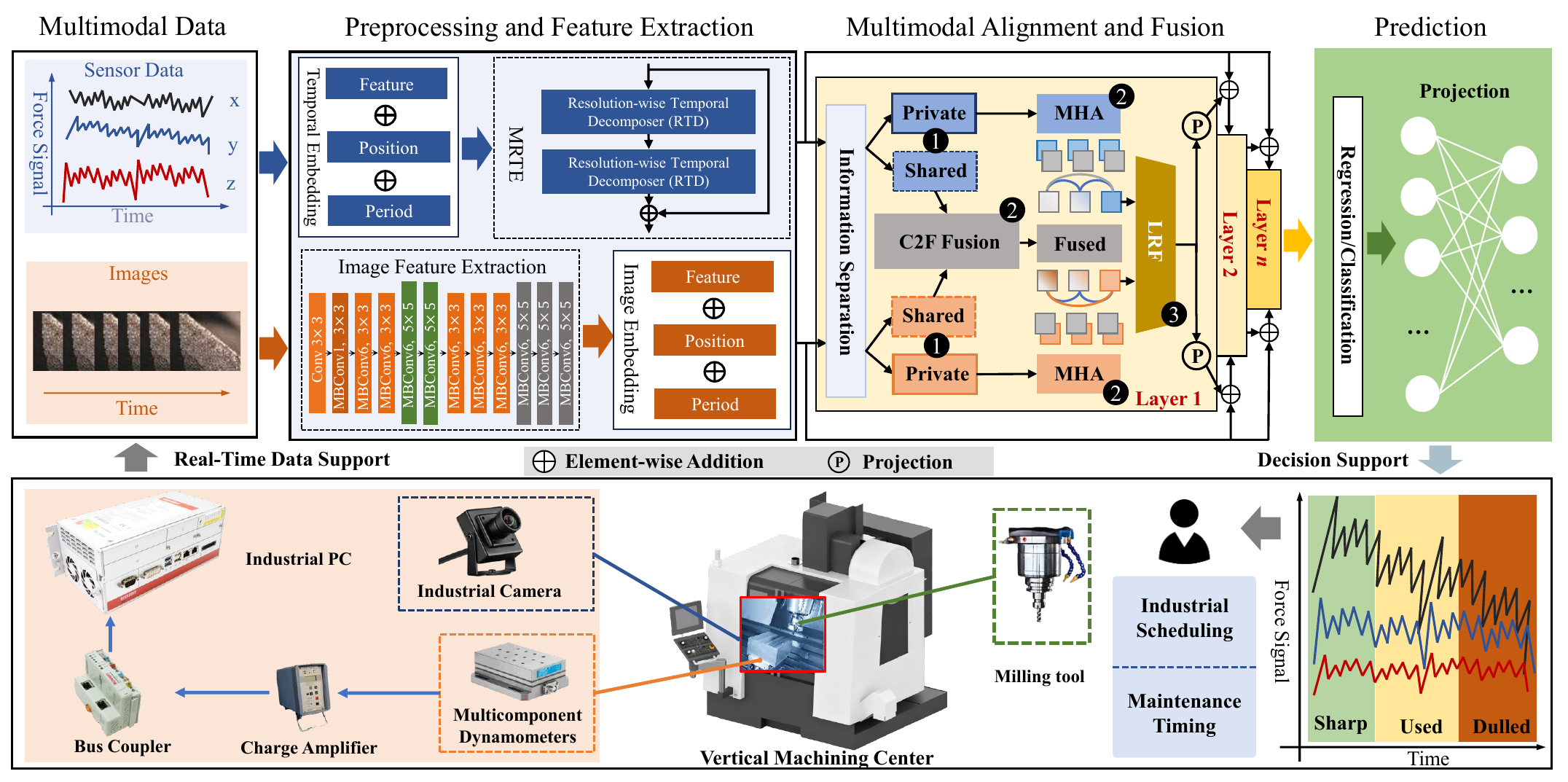}
\caption{Overall architecture of \textsc{OmniFuser}.  
Temporal and spatial features are first extracted through dedicated modules and then progressively fused by the $\text{C}^2\text{F}$ module. 
The resulting fused representation is used for downstream prediction tasks.}
\label{overall}
\end{figure*}

The overall architecture of \textsc{OmniFuser}, illustrated in Fig.~\ref{overall}, is structured as an end-to-end predictive maintenance service comprising three stages: Preprocessing and Feature Extraction, Multimodal Alignment and Fusion, and Prediction. Sensor signals and tool images are first encoded into temporal and spatial features, which are then embedded into a shared representation space. The core $\text{C}^2\text{F}$ (Contamination-free Cross-modal Fusion) module subsequently performs three key operations: (i) separating shared and private components per modality, (ii) applying Proxy-based Cross-Modal Attention (PCMA) on shared features with learnable landmark keys while refining private features via self-attention, and (iii) fusing the enhanced representations through a learnable hybrid gate that combines global and local gating effects. The resulting fused representation enables the service to simultaneously forecast future force signals and classify tool wear states. In our design, each component of $\text{C}^2\text{F}$, i.e., cross-modal fusion, adaptive gating, and recursive refinement, is designed as a loosely coupled, reusable service module. This facilitates integration into existing industrial IoT service pipelines (e.g., via RESTful APIs or edge microservices), supporting dynamic composition with other analytics services such as anomaly detection or remaining useful life estimation. Main notations are listed in Table~\ref{Notations_C}, and the functional roles of each module are summarized in Table~\ref{module}.

\begin{table}[!ht]
\centering
\scriptsize
\caption{Main Notations}
\label{Notations_C}
\begin{tabular}{@{}ll@{}}
\toprule
\textbf{Notation} & \textbf{Definition} \\
\midrule
$\mathbf{X}$ & Multidimensional sensor signal reflecting various physical features \\
$\mathbf{I}$ & Image sequence capturing tool surface after each milling operation \\
$\mathbf{H}^\text{r}$ & Embedded sensor signal \\
$\mathbf{Z}^\text{r}$ and $\mathbf{Z}^\text{i}$ & Sensor and image features after feature extraction \\
$\textrm{S}^m$ and $\textrm{P}^m$ & Shared and private features of each modality \\
$\mathbf{\hat{P}}^m$ & Enhanced intra-modality feature \\
$\mathbf{H}^{\textrm{r}\leftrightarrow \textrm{i}}$ & Enhanced mutual information cross-correlation \\
$\mathbf{G}^\textrm{m}$ & Learnable hybrid gate \\
$\hat{\mathbf{Z}}$ & Fused feature after low-rank fusion \\
$\hat{\mathbf{Z}}^{\textrm{(n)}}$ & Final fused feature after recursive refinement \\
\bottomrule
\end{tabular}
\end{table}

\begin{table*}[htbp]
\scriptsize
\centering
\caption{Module Descriptions}
\label{module}
\begin{tabular}{lll}
\toprule
Method & Key Characteristics \\
\midrule
$\text{C}^2\text{F}$ (Contamination-free Cross-modal Fusion) & 
Multimodal fusion strategy that enables contamination-free feature alignment across modalities \\

PCMA (Proxy-based Cross-Modal Attention) & 
Adaptive landmark selection for efficient cross-modal fusion \\

MRTE (Multi-Resolution Temporal Extractor) & 
Temporal feature extractor with multiple resolutions, enhancing robustness to long- and short-range dependencies\\

RTD (Resolution-wise Temporal Decomposer) & 
Resolution-wise low-rank decomposition for interpretable dynamics \\
\bottomrule
\end{tabular}
\end{table*}


\subsection{Preprocessing and Feature Extraction}
Sensor signals $\mathbf{X}$ are typically low-dimensional but rich in dynamic patterns, whereas image data $\mathbf{I}$ is high-dimensional and semantically sparse. This fundamental difference motivates a modality-specific preprocessing strategy: sensor data is first embedded into a high-dimensional latent space to facilitate flexible temporal modeling, while image data undergoes feature extraction to reduce redundancy and noise before introducing temporal structure for alignment.

For each sensor signal $\mathbf{x}_t$, it first passes through a multi-layer perceptron (MLP) to generate value embeddings: $\mathbf{E}^\text{val}_t = \textrm{MLP}(\mathbf{x}_t) \in \mathbb{R}^d$. This transformation maps low-dimensional inputs into a unified latent space, enabling nonlinear feature interaction that captures their distinct temporal patterns. The MLP consists of two stacked linear layers, with the final layer projecting onto dimension $d$. To incorporate sequential order, a fixed positional embedding is added, computed using sine and cosine functions with predefined frequencies: $\mathbf{E}^\text{pos}_t = \textrm{PE}[t] \in \mathbb{R}^d$, where $\textrm{PE}[t]$ denotes the $t$-th row of a deterministic positional encoding matrix. Additionally, high-level temporal periodicity is encoded using learnable embeddings for discrete time attributes. For each time step $t$, the period embedding is given by:
\begin{equation}
\mathbf{E}^\text{per}_t = \textrm{TE}_\text{c}(\textrm{c}_t) + \textrm{TE}_\text{b}(\textrm{b}_t) + \textrm{TE}_\text{p}(\textrm{p}_t) \in \mathbb{R}^d,
\end{equation}
where $\textrm{c}_t$, $\textrm{b}_t$, and $\textrm{p}_t$ represent spindle cycle, cutting batch, and relative position within the cycle at time step $t$, respectively, and $\textrm{TE}_*(\cdot)$ denotes a learnable embedding lookup table for each attribute. Summing these embeddings yields an enriched representation $\mathbf{H}^\text{r} \in \mathbb{R}^{T \times d}$.



\begin{figure}    
[!ht]
\centering
\includegraphics[width=0.8\linewidth]{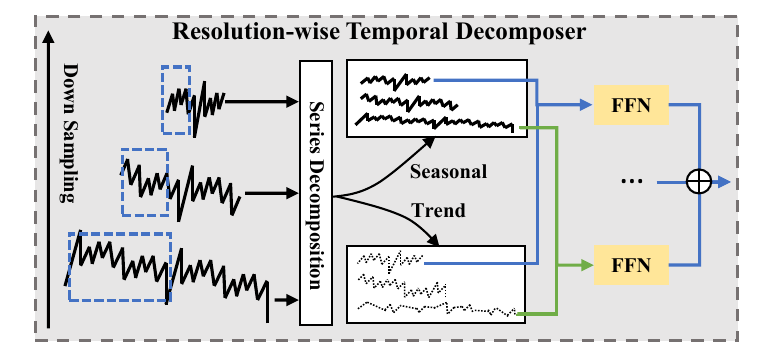}
\caption{The architecture of RTD.}
\label{RTD}
\end{figure}

Next, we extract temporal structures using Multi-Resolution Temporal Extractor (MRTE), designed to model temporal dependencies at various resolutions. Tool degradation exhibits multiperiodic temporal patterns (Fig. \ref{mo1}): long-term wear progresses gradually across machining passes, while short-term fluctuations are driven by spindle motion and vibrations. MRTE addresses this by stacking several Resolution-wise Temporal Decomposer (RTD) blocks in a residual manner, as shown in Fig. \ref{RTD}. Within RTD, the input $\mathbf{H}^\text{r}$ is downsampled using depthwise temporal convolution with stride $s_l$. Let $T_l = \lceil T / s_l \rceil$ denote the temporal length after downsampling. The resulting representation is given by:
\begin{equation}
    \hat{\mathbf{H}}^{(l)}_{t} 
    = \left( \sum_{\tau=0}^{k_l-1} 
    \mathbf{H}^{\textrm{r}}_{\,t s_l + \tau} \odot \mathbf{k}^{(l)}_{\tau} \right)\mathbf{W}^{(l)},
    \quad t = 0,\dots,T_l-1,
\end{equation}
where $\mathbf{k}^{(l)}_{\tau} \in \mathbb{R}^{d}$ represents the $\tau$-th depthwise convolution weight vector, $\odot$ denotes the Hadamard product, and $\mathbf{W}^{(l)} \in \mathbb{R}^{d \times d}$ is the channel projection matrix. The kernel size $k_l$ equals the stride $s_l$, and zero padding is applied when necessary to maintain consistent temporal coverage.

The downsampled sequence captures dynamics at different resolutions, enabling coarse scales to reflect long-term wear trends and fine scales to preserve short-term periodic fluctuations. These sequences are decomposed into trend and seasonal components through a learnable low-rank projection:
\begin{equation}
    \mathbf{T}^{(l)} = \hat{\mathbf{H}}^{(l)} \mathbf{P}^{(l)} \mathbf{Q}^{(l)}, 
    \qquad 
    \mathbf{S}^{(l)} = \hat{\mathbf{H}}^{(l)} - \mathbf{T}^{(l)},
\end{equation}
where $\mathbf{P}^{(l)} \in \mathbb{R}^{d \times r}$ and $\mathbf{Q}^{(l)} \in \mathbb{R}^{r \times d}$ are learnable projection matrices. After decomposition, the two components are recombined within the same temporal length $T_l$: $\mathbf{U}^{(l)} = \mathbf{T}^{(l)} + \mathbf{S}^{(l)}$. Each $\mathbf{U}^{(l)}$ is then processed through an independent feed-forward module that restores the temporal length to $T$: $\tilde{\mathbf{U}}^{(l)} = \textrm{FFD}^{(l)}(\mathbf{U}^{(l)}) \in \mathbb{R}^{T \times d}$. Finally, outputs from multiple resolutions are fused to form the unified temporal representation: $\mathbf{Z}^{\textrm{r}} = \sum_{l=1}^{L} \tilde{\mathbf{U}}^{(l)}$. 

In parallel, image frame $\mathbf{I}_t$ is processed by a lightweight EfficientNet backbone to extract spatial features:
\begin{equation}
    \mathbf{f}_t = \textrm{EffNet}(\mathbf{I}_t) \in \mathbb{R}^{d_\text{raw}}.
\end{equation}
To capture temporal patterns, a graph attention network (GAT) is applied over the extracted image feature sequences, yielding image value embeddings: $\bar{\mathbf{E}}^\text{val}_t = [\textrm{GAT}([\mathbf{f}_1, \dots, \mathbf{f}_T])]_t \in \mathbb{R}^d$. Each frame is treated as a node in a fully connected temporal graph, with edge weights adaptively learned to reflect cross-frame relevance. Positional and periodic embeddings are then added to the image value embeddings, producing $\mathbf{Z}^\text{i}$.

\subsection{Multimodal Alignment and Fusion}
\begin{figure}    
[!ht]
\centering
\includegraphics[width=0.74\linewidth]{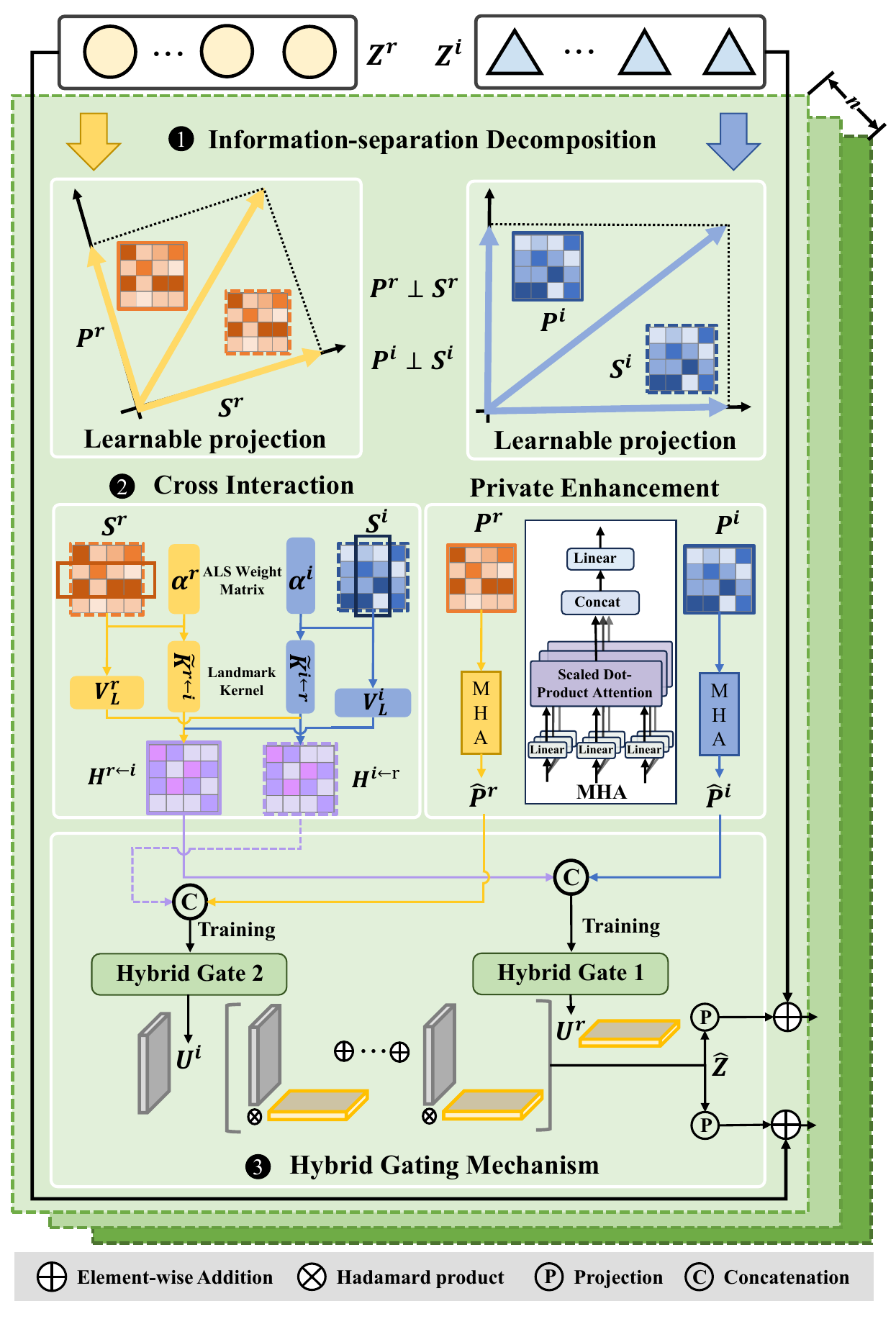}
\caption{Illustration of the $\text{C}^2\text{F}$ architecture.}
\label{c2f}
\end{figure}

$\text{C}^2\text{F}$ is designed to generate an expressive joint representation from $\mathbf{Z}^\textrm{r}$ and $\mathbf{Z}^\textrm{i}$ through a principled three-stage process (Fig.~\ref{c2f}).  
\circlednum{1} It first performs information-separation decomposition, explicitly projecting each modality into shared and modality-specific components. This prevents cross-modal information contamination and preserves modality-unique cues, enabling subsequent learning to focus explicitly on complementary patterns rather than redundant correlations.  
\circlednum{2} It then conducts efficient cross-modal interaction on the shared components via a PCMA mechanism, which leverages key landmark representations to capture cross-modal dependencies with significantly reduced computational complexity.  
\circlednum{3} Finally, it adaptively combines the interacted shared features with the private (modality-specific) features using a hybrid gating mechanism that unifies global responses and local patterns, followed by a projection that compresses the fused representation into a compact, task-oriented form. 

\circlednum{1} The first step of $\text{C}^2\text{F}$ explicitly disentangles the input representations $\mathbf{Z}^\textrm{r}$ and $\mathbf{Z}^\textrm{i}$ into modality-specific and shared components.  
For each modality $m \in \{\textrm{r}, \textrm{i}\}$, a projection matrix $\mathbf{W}'_m \in \mathbb{R}^{d \times d}$ is learned to map the input features into a shared subspace: $\mathbf{S}^m = \mathbf{Z}^m \mathbf{W}'_m$. The modality-private component is then obtained by orthogonally projecting the original features onto the complement of this shared subspace:
\begin{equation}
\mathbf{P}^m = \mathbf{Z}^m - \frac{\langle \mathbf{Z}^m, \mathbf{S}^m \rangle_F}{\|\mathbf{S}^m\|_F^2 + \varepsilon}\mathbf{S}^m,
\label{eq:private-proj}
\end{equation}
where $\langle \mathbf{A}, \mathbf{B} \rangle_F = \textrm{Tr}(\mathbf{A}^\textrm{T} \mathbf{B})$ denotes the Frobenius inner product, $\|\mathbf{S}^m\|_F$ is the Frobenius norm, and $\varepsilon$ is a small constant ensuring numerical stability. Consequently, each modality representation is decomposed as $\mathbf{Z}^m = \mathbf{P}^m + \mathbf{S}^m$, with $\mathbf{P}^m \perp \mathbf{S}^m$. This orthogonal decomposition satisfies $I(\mathbf{Z}^\textrm{r}; \mathbf{Z}^\textrm{i}) \geq I(\mathbf{S}^\textrm{r}; \mathbf{S}^\textrm{i})$, 
where $I(\cdot;\cdot)$ denotes mutual information, quantifying the statistical dependency between two random variables. This inequality indicates that the mutual information between the original representations $\mathbf{Z}^\textrm{r}$ and $\mathbf{Z}^\textrm{i}$ is lower bounded by that of their shared components $\mathbf{S}^\textrm{r}$ and $\mathbf{S}^\textrm{i}$. Thus, the decomposition preserves cross-modal dependencies, ensuring that the shared representation retains the information necessary for subsequent fusion.

\begin{mytheorem}
Given the modality-specific decomposition $\mathbf{Z}^m = \mathbf{P}^m + \mathbf{S}^m$ with $\mathbf{P}^m \perp \mathbf{S}^m$ for modality $m \in \{\textrm{r}, \textrm{i}\}$, the mutual information between the original modalities is lower bounded by that of the shared components: 
\begin{equation}
    I(\mathbf{Z}^\textrm{r}; \mathbf{Z}^\textrm{i}) \geq I(\mathbf{S}^\textrm{r}; \mathbf{S}^\textrm{i}).
\end{equation}
\label{thm1}
\end{mytheorem}

\begin{proof}
The shared component $\mathbf{S}^m$ is obtained from $\mathbf{Z}^m$ via a deterministic projection onto a learned shared subspace. There exists a mapping $\boldsymbol{\Pi}_m$ such that
\begin{equation}
    \mathbf{S}^m = \boldsymbol{\Pi}_m(\mathbf{Z}^m), 
    \qquad 
    \mathbf{P}^m = \mathbf{Z}^m - \boldsymbol{\Pi}_m(\mathbf{Z}^m),
\end{equation}
which ensures $\mathbf{P}^m \perp \mathbf{S}^m$ through the Frobenius-orthogonal projection in \eqref{eq:private-proj}.  
Since $\mathbf{S}^m$ is a deterministic function of $\mathbf{Z}^m$, the data processing inequality for mutual information applies \cite{inequality}: for any random variables $X, Y$ and deterministic mappings $f, g$,
\begin{equation}
    I(X; Y) \ge I(f(X); Y) \ge I(f(X); g(Y)).
\end{equation}
Setting $X = \mathbf{Z}^\textrm{r}$, $Y = \mathbf{Z}^\textrm{i}$, $f = \boldsymbol{\Pi}_\textrm{r}$, and $g = \boldsymbol{\Pi}_\textrm{i}$ yields
\begin{equation}
    I(\mathbf{Z}^\textrm{r}; \mathbf{Z}^\textrm{i})
\ge I(\mathbf{S}^\textrm{r}; \mathbf{Z}^\textrm{i})
\ge I(\mathbf{S}^\textrm{r}; \mathbf{S}^\textrm{i}),
\end{equation}
which proves the claim.
\end{proof}
Theorem~\ref{thm1} ensures that the orthogonal decomposition retains all necessary cross-modal dependencies in the shared subspace, providing a contamination-free and information-sufficient basis for fusion.


\circlednum{2} The second step of $\text{C}^2\text{F}$ processes the decomposed features through two parallel pathways: a modality-private self-attention path and a shared cross-modal interaction path. For each modality $m \in \{\textrm{r}, \textrm{i}\}$, the private component $\mathbf{P}^m$ is refined by an intra-modality multi-head self-attention (MHA) module: $\mathbf{\hat{P}}^m = \textrm{MHA}(\mathbf{P}^m)$, which enhances long-range temporal dependencies unique to the modality while suppressing irrelevant variations. 

Concurrently, the shared components $\mathbf{S}^\textrm{r}$ and $\mathbf{S}^\textrm{i}$ are processed by the PCMA mechanism. To avoid the quadratic complexity of full attention, PCMA approximates the cross-modal attention map using a compact set of adaptive landmarks. Specifically, we introduce an Adaptive Landmark Selection (ALS) mechanism that dynamically constructs $k$ landmarks from each input sequence. For modality $m$, each token $\mathbf{S}^m_t$ is assigned a soft weight vector $\alpha^m_t = \textrm{softmax}\big(f^m(\mathbf{S}^m_t)\big) \in \mathbb{R}^k$, where $f^m: \mathbb{R}^d \to \mathbb{R}^k$ is a lightweight scoring function. Stacking $\{\alpha^m_t\}_{t=1}^T$ row-wise yields $\alpha^m \in \mathbb{R}^{T \times k}$, and the landmarks are obtained via weighted aggregation:
\begin{equation}
\mathbf{L}^m = (\alpha^m)^\textrm{T} \mathbf{S}^m, \qquad \mathbf{L}^m \in \mathbb{R}^{k \times d}, \quad k \ll T.
\end{equation}
Cross-modal messages are then computed using the other modality’s landmarks:
\begin{equation}
\mathbf{H}^{\textrm{r}\leftarrow \textrm{i}} = \rho\!\left(\frac{\mathbf{S}^{\textrm{r}}(\mathbf{L}^{\textrm{i}})^\textrm{T}}{\sqrt{d}}\right)\mathbf{V}^{\textrm{i}}_{\mathbf{L}}, \quad
\mathbf{H}^{\textrm{i}\leftarrow \textrm{r}} = \rho\!\left(\frac{\mathbf{S}^{\textrm{i}}(\mathbf{L}^{\textrm{r}})^\textrm{T}}{\sqrt{d}}\right)\mathbf{V}^{\textrm{r}}_{\mathbf{L}},
\end{equation}
where $\rho$ denotes row-wise softmax. The landmark-space value matrices are constructed analogously: $\mathbf{V}^m = \mathbf{S}^m \mathbf{W}^m_v$ and $\mathbf{V}^m_{\mathbf{L}} = (\alpha^m)^\textrm{T} \mathbf{V}^m$, with $\mathbf{W}^m_v \in \mathbb{R}^{d \times d}$. This design reduces the computational complexity from $\mathcal{O}(T^2 d)$ to $\mathcal{O}(T k d)$ while preserving dominant cross-modal dependencies. 

\begin{mytheorem}
Let $\mathbf{K} = \rho\big(\mathbf{S}^\textrm{r}(\mathbf{S}^\textrm{i})^\textrm{T} / \sqrt{d}\big)$ denote the cross-modal attention kernel, and let $\tilde{\mathbf{K}}$ be its PCMA-based approximation using $k$ adaptive landmarks per modality. Then the approximated attention output $\tilde{\mathbf{H}}^{\textrm{r}\leftarrow \textrm{i}} = \tilde{\mathbf{K}} \mathbf{V}^\textrm{i}$ satisfies
\begin{equation}
\mathbb{E}\Big[\big\|\mathbf{H}^{\textrm{r}\leftarrow \textrm{i}} - \tilde{\mathbf{H}}^{\textrm{r}\leftarrow \textrm{i}}\big\|_F\Big]
\le 
c\,\sqrt{\frac{T}{k}}\,
\|\mathbf{K}\|_F\,
\|\mathbf{V}^{\textrm{i}}\|_F,
\end{equation}
where $\mathbb{E}[\cdot]$ is taken over the data-dependent landmark selection, $c$ is a constant independent of $T$, $k$, and $d$, and $\|\cdot\|_F$ denotes the Frobenius norm.
\end{mytheorem}

\begin{proof}
Let $\mathbf{H}^{\textrm{r}\leftarrow \textrm{i}} = \mathbf{K} \mathbf{V}^{\textrm{i}}$ and $\tilde{\mathbf{H}}^{\textrm{r}\leftarrow \textrm{i}} = \tilde{\mathbf{K}} \mathbf{V}^{\textrm{i}}$, where $\tilde{\mathbf{K}}$ is the PCMA approximation of $\mathbf{K}$. The error is $\mathbf{H}^{\textrm{r}\leftarrow \textrm{i}} - \tilde{\mathbf{H}}^{\textrm{r}\leftarrow \textrm{i}} = (\mathbf{K} - \tilde{\mathbf{K}}) \mathbf{V}^{\textrm{i}}$. By submultiplicativity of the Frobenius norm,
\begin{equation}
\big\| \mathbf{H}^{\textrm{r}\leftarrow \textrm{i}} - \tilde{\mathbf{H}}^{\textrm{r}\leftarrow \textrm{i}} \big\|_F
\leq \|\mathbf{K} - \tilde{\mathbf{K}}\|_F \, \|\mathbf{V}^{\textrm{i}}\|_F.
\label{eq:bound1}
\end{equation}
The PCMA approximation constructs $\tilde{\mathbf{K}}$ using adaptive landmarks $\mathbf{L}^{\textrm{r}} = (\alpha^{\textrm{r}})^\textrm{T} \mathbf{S}^{\textrm{r}}$ and $\mathbf{L}^{\textrm{i}} = (\alpha^{\textrm{i}})^\textrm{T} \mathbf{S}^{\textrm{i}}$, which induces a CUR-type decomposition $\tilde{\mathbf{K}} = \mathbf{C} \mathbf{U} \mathbf{R}$, where $\mathbf{C}$ and $\mathbf{R}$ are weighted subsets of columns and rows of $\mathbf{K}$ selected via the ALS weights $\alpha^{\textrm{i}}$ and $\alpha^{\textrm{r}}$.  

Since attention kernels are typically near low-rank due to strong cross-modal correlations, and ALS is trained end-to-end to emphasize informative tokens, the selected landmarks effectively span the dominant subspace of $\mathbf{K}$. Under this condition, standard CUR approximation theory~\cite{CUR} yields
\begin{equation}
\mathbb{E}\big[ \|\mathbf{K} - \tilde{\mathbf{K}}\|_F \big] \leq c \sqrt{\frac{T}{k}} \, \|\mathbf{K}\|_F,
\label{eq:bound2}
\end{equation}
where the expectation accounts for data-dependent landmark selection, and $c$ is a constant independent of $T$, $k$, and $d$. Substituting \eqref{eq:bound2} into \eqref{eq:bound1} gives the desired result:
\begin{equation}
    \mathbb{E}\Big[ \big\| \mathbf{H}^{\textrm{r}\leftarrow \textrm{i}} - \tilde{\mathbf{H}}^{\textrm{r}\leftarrow \textrm{i}} \big\|_F \Big]
\leq c \sqrt{\frac{T}{k}} \, \|\mathbf{K}\|_F \, \|\mathbf{V}^{\textrm{i}}\|_F.
\end{equation}
\end{proof}

This bound guarantees that PCMA preserves cross-modal interaction fidelity with provably controlled error, while reducing complexity from $\mathcal{O}(T^2 d)$ to $\mathcal{O}(T k d)$. The outputs of this stage are the refined private features $\mathbf{\hat{P}}^\textrm{r}, \mathbf{\hat{P}}^\textrm{i}$ and the cross-attended shared features $\mathbf{H}^{\textrm{r}\leftarrow \textrm{i}}, \mathbf{H}^{\textrm{i}\leftarrow \textrm{r}}$, which jointly encode modality-specific dynamics and cross-modal complementary information.

\circlednum{3} The final step of $\text{C}^2\text{F}$ fuses the outputs from the private and shared pathways. For each modality, the cross-attended shared features and refined private features are concatenated along the feature dimension:
\begin{equation}
    \mathbf{F}^\textrm{r} = \big[\mathbf{H}^{\textrm{r}\leftarrow \textrm{i}},\, \hat{\mathbf{P}}^\textrm{r}\big], \quad
    \mathbf{F}^\textrm{i} = \big[\mathbf{H}^{\textrm{i}\leftarrow \textrm{r}},\, \hat{\mathbf{P}}^\textrm{i}\big].
\end{equation}

To adaptively balance global degradation trends and local temporal fluctuations, we introduce a hybrid gating mechanism. Taking the sensor modality as an example, the gate is computed as
\begin{equation}
\mathbf{G}^\textrm{r} = \alpha \cdot \sigma\big(\mathbf{W}_\textrm{l} \mathbf{F}^\textrm{r}\big) + (1 - \alpha) \cdot \sigma\big(\textrm{Conv1d}(\mathbf{F}^\textrm{r})\big),
\end{equation}
where $\mathbf{W}_\textrm{l} \in \mathbb{R}^{2d \times 2d}$ captures global dependencies via a linear projection, $\textrm{Conv1d}(\cdot)$ is a depthwise 1D convolution modeling local temporal patterns, $\sigma(\cdot)$ is the sigmoid function, and $\alpha \in [0,1]$ is a learnable scalar weight. The gated representation is then formed by selectively blending shared and private streams:
\begin{equation}
\mathbf{U}^\textrm{r} = \mathbf{G}^\textrm{r} \odot \mathbf{H}^{\textrm{r}\leftarrow \textrm{i}} + (1 - \mathbf{G}^\textrm{r}) \odot \hat{\mathbf{P}}^\textrm{r},
\end{equation}
with $\mathbf{U}^\textrm{i}$ defined symmetrically for the image modality. The modality-specific gated features $\mathbf{U}^\textrm{r}$ and $\mathbf{U}^\textrm{i}$ are concatenated and projected through a low-rank mapping to yield the final fused representation:
\[
\hat{\mathbf{Z}} = \textrm{LR}\big([\mathbf{U}^\textrm{r}, \mathbf{U}^\textrm{i}]\big),
\]
where $\textrm{LR}(\cdot)$ denotes a two-layer bottleneck projection with hidden dimension $r \ll 2d$:
\begin{equation}
\textrm{LR}(\mathbf{x}) = \mathbf{W}_3\, \phi(\mathbf{W}_2 \mathbf{x}), \quad
\mathbf{W}_2 \in \mathbb{R}^{2d \times r},\;
\mathbf{W}_3 \in \mathbb{R}^{r \times d},
\end{equation}
and $\phi(\cdot)$ is the GELU activation. This design achieves efficient fusion with reduced computational overhead while preserving task-relevant discriminative capacity.

To prevent irreversible information loss from a single-pass fusion, particularly for subtle or gradually evolving wear patterns, we further employ a recursive refinement strategy that anchors the fusion process to the original inputs. Let $\mathbf{Z}^{\textrm{ori}} = \{\mathbf{Z}^{\textrm{r}}, \mathbf{Z}^{\textrm{i}}\}$ denote the original multimodal features. Starting from $\hat{\mathbf{Z}}^{(1)} = \mathcal{G}(\mathbf{Z}^{\textrm{r}}, \mathbf{Z}^{\textrm{i}})$, the $r$-th refinement step ($r = 2, \dots, n$) updates the fused representation as
\begin{equation}
\hat{\mathbf{Z}}^{(r)} = \mathcal{G}\Big( \mathbf{Z}^{\textrm{r}} + \mathcal{P}_\textrm{r}\big(\hat{\mathbf{Z}}^{(r-1)}\big),\;
\mathbf{Z}^{\textrm{i}} + \mathcal{P}_\textrm{i}\big(\hat{\mathbf{Z}}^{(r-1)}\big) \Big),
\end{equation}
where $\mathcal{P}_\textrm{r}(\cdot)$ and $\mathcal{P}_\textrm{i}(\cdot)$ are lightweight modality-specific projection layers, and $\mathcal{G}(\cdot)$ denotes the full $\text{C}^2\text{F}$ module. This residual-style re-injection of original features mitigates representation drift and enhances sensitivity to long-term degradation dynamics in milling operations.

\subsection{Milling Tool Condition Forecast}
Let the fused multimodal representation over the observation window be denoted as $\hat{\mathbf{Z}} = [\hat{\mathbf{z}}_{1}, \ldots, \hat{\mathbf{z}}_{T}] \in \mathbb{R}^{T \times d}$. For future cutting force prediction, we map $\hat{\mathbf{Z}}$ to a continuous multivariate output sequence via a regression function $f_{\textrm{reg}}(\cdot)$. Specifically, at each future time step $t = T+1, \dots, T+P$, the predicted force vector is computed using only the final fused feature $\hat{\mathbf{z}}_T$:
\begin{equation}
    \hat{\mathbf{y}}_{t} = f_{\textrm{reg}}(\hat{\mathbf{Z}}, t)
    = \mathbf{W}_{r}\hat{\mathbf{z}}_{T} + \mathbf{b}_{r},
\end{equation}
where $\mathbf{W}_{r} \in \mathbb{R}^{d \times C}$ and $\mathbf{b}_{r} \in \mathbb{R}^{C}$ are learnable parameters. This yields the predicted force sequence $\hat{\mathbf{Y}} = [\hat{\mathbf{y}}_{T+1}, \ldots, \hat{\mathbf{y}}_{T+P}] \in \mathbb{R}^{P \times C}$.

For tool wear classification, the same fused representation is processed by a classification head $f_{\textrm{cls}}(\cdot)$. At each future step $t$, categorical logits are generated from $\hat{\mathbf{z}}_T$:
\begin{equation}
    \mathbf{o}_{t} = f_{\textrm{cls}}(\hat{\mathbf{z}}_{T}) \in \mathbb{R}^{3},
    \quad t = T+1,\ldots,T+P.
\end{equation}
Applying the softmax function converts logits to class probabilities over the three wear states $\{\text{Sharp}, \text{Used}, \text{Dulled}\}$. The predicted label at time $t$ is obtained by
\begin{equation}
    \hat{s}_{t} = \arg\max_{k} \textrm{Softmax}(\mathbf{o}_{t})_k,
\end{equation}
resulting in the wear state sequence $\hat{\mathbf{S}} = [\hat{s}_{T+1}, \ldots, \hat{s}_{T+P}]$.

\section{Performance Evaluation}
\label{PE}
\subsection{Experimental Setup}

\subsubsection{Datasets and Synchronization Strategy}

We conduct experiments on two real-world multimodal tool-wear datasets: MATWI~\cite{MATWI} and Mudestreda~\cite{Mudestreda}. 
\begin{itemize}
    \item The MATWI dataset is collected on a standardized milling test bench equipped with a triaxial force dynamometer and an industrial camera. It provides synchronized cutting-force signals at \(\text{500 Hz}\) and high-resolution tool images (3072 $\times$ 2048~px) captured after each cutting pass. In total, MATWI contains 720 recorded sequences covering complete wear trajectories of six tools. As wear categories are not annotated, this dataset is mainly employed for forecasting rather than classification.
    \item The Mudestreda dataset extends the experimental setting to realistic industrial conditions with varying spindle speeds (400-1200~rpm), feed rates (0.05-0.3~mm/rev), and materials (aluminum, steel, titanium). 
    It includes both force and image modalities, but with more complex degradation behaviors and cross-condition variability. Each data point is annotated with discrete wear states (\textit{Sharp, Used, Dulled}) by expert inspection, which enables both tool-state classification and force signal forecasting tasks to be conducted on this dataset.
    Mudestreda exhibits more complex degradation behaviors and cross-condition variability. The dataset further covers a wide range of machining scenarios, including dry and wet cutting conditions, variable illumination, sensor noise, and coolant interference, ensuring that both visual and signal modalities capture realistic disturbances encountered in production lines.
\end{itemize}
Across the two datasets, the experiments collectively span diverse operating regimes and tool geometries, covering most practical milling conditions encountered in small- and medium-scale manufacturing.

Tool images are captured only once per cutting pass, while sensor signals are sampled at high frequency. To align the two modalities, we replicate each image across all sensor readings within the same pass, which is justified by the fact that tool wear changes slowly and remains nearly constant during a single pass. Replication is limited to within-pass to avoid overfitting, and updated images are used in subsequent passes to reflect actual wear progression. A control experiment (halving image update frequency) showed negligible performance change (±1.5\% accuracy), confirming that this strategy introduces no significant bias. The replicated images thus serve as stable visual anchors that complement the fast-varying sensor dynamics. Our alignment ensures compatibility with such asynchronous multimodal service interfaces without requiring costly hardware synchronization.


\subsubsection{Baselines and Evaluation Metrics}
We evaluate \textsc{OmniFuser} against three representative categories of baselines.  
First, we consider time series forecasting models tailored for predictive maintenance. TimeKAN~\cite{TimeKAN} and FilterTS~\cite{FilterTS} leverage frequency-domain decomposition to jointly capture long-term degradation trends and short-term fluctuations. TimePFN~\cite{TimePFN} is specifically designed for scenarios with scarce labels. MSGNet~\cite{MSGNet}, TimeMixer~\cite{TimeMixer}, and FEDformer~\cite{FEDformer} explicitly model trend-seasonality structures that align well with the progressive nature of tool wear. Meanwhile, iTransformer~\cite{iTransformer} and TimesNet~\cite{TimesNet} enhance inter-variable interactions and periodicity modeling, both of which are crucial for capturing the coupled dynamics among multiple sensor streams.
Second, we include established multimodal fusion architectures: CDA~\cite{CDA}, MBT~\cite{MBT}, LMF~\cite{LMF}, and TFN~\cite{TFN}. To ensure a fair comparison of fusion strategies, each is integrated into our framework by replacing only the $\text{C}^2\text{F}$ module while keeping all other components and training settings identical.
Third, we benchmark against recent multimodal large forecasting models, including Chronos~\cite{Chronos}, TimesFM~\cite{TimesFM}, Moirai~\cite{Moirai}, and Lag-Llama~\cite{Lag-Llama}, which represent the state of the art in foundation-model-based time series prediction. Their evaluation on tool wear forecasting reveals how effectively general temporal priors can be adapted to domain-specific predictive maintenance tasks.
All experiments are conducted on an NVIDIA GeForce RTX 4090 GPU with 24~GB memory.

For performance assessment, we adopt two sets of evaluation metrics corresponding to the two tasks. In forecasting~\cite{accuracy}, we use Mean Squared Error (MSE) and Mean Absolute Error (MAE). MSE is sensitive to large deviations and thus emphasizes sharp fluctuations in cutting force signals, whereas MAE provides a robust measure of average prediction error that is less affected by outliers. In both cases, lower values indicate better predictive accuracy. For classification~\cite{class}, we report Accuracy, Precision, Recall, and F1-Score. Together, these metrics reflect overall correctness, control over false positives, sensitivity to degraded tool states, and their balanced trade-off. Higher values indicate stronger discriminative performance.

\subsubsection{Parameter Sensitivity}
\begin{figure*}[!t]
\centering
\begin{minipage}{0.49\textwidth}
    \centering
    \includegraphics[width=\linewidth]{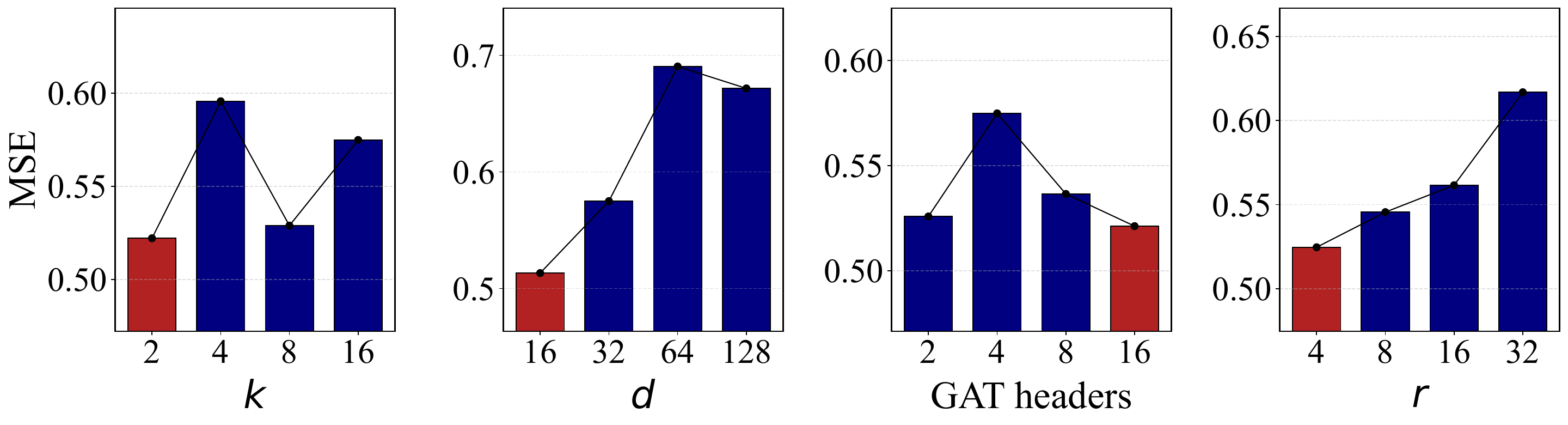}
    \caption{Parameter Sensitivity Analysis of parameters including $k$, $d$, GAT heads and $r$.}
    \label{abalation_mse1}
\end{minipage}
\hfill
\begin{minipage}{0.49\textwidth}
    \centering
    \includegraphics[width=\linewidth]{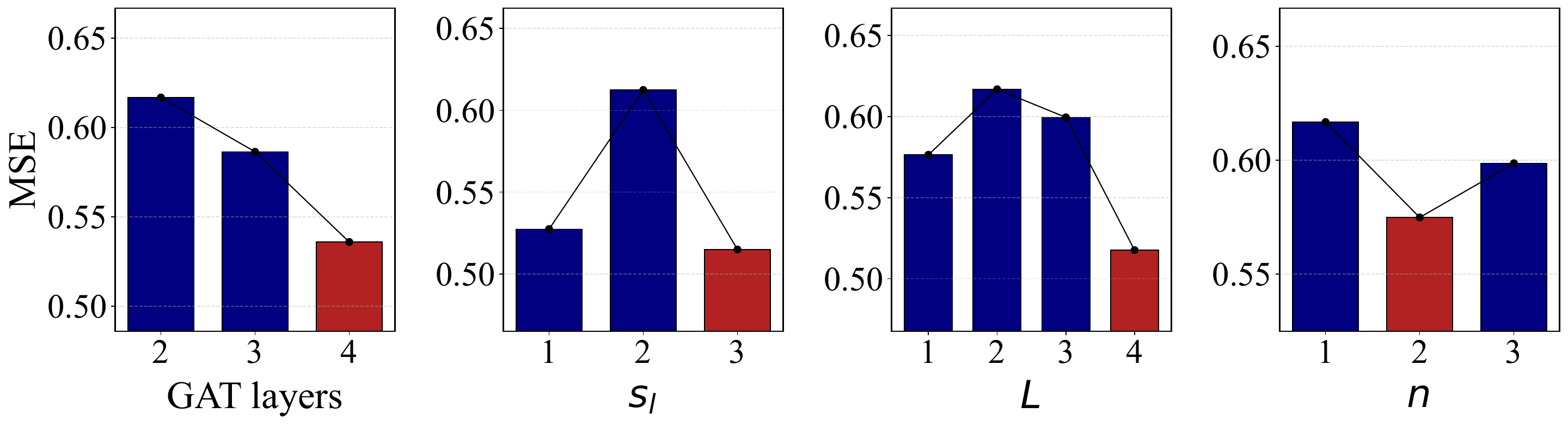}
    \caption{Parameter Sensitivity Analysis of parameters including GAT layers, $s_l$, $L$ and $n$.}
    \label{abalation_mse2}
\end{minipage}
\end{figure*}

In our implementation, several hyperparameters are identified as critical to predictive performance. A subset of these is designated as adjustable and further analyzed through sensitivity studies on the Mudestreda dataset (see Figs.~\ref{abalation_mse1} and~\ref{abalation_mse2}). Specifically, we tune the following: The embedding dimension ($d = 32$) balances representation capacity against computational overhead. The number of MRTE layers ($L = 4$) controls the depth of multiresolution temporal modeling. The downsampling stride ($s_l = 3$) governs the trade-off between temporal granularity and computational efficiency. The projection rank ($r = 8$) enables compact yet expressive trend-seasonal decomposition. For visual modeling, the GAT is configured with four layers and four attention heads to strengthen cross-modal relational learning. Within the $\text{C}^2\text{F}$ module, the number of landmarks $k$ is set to 4 to capture salient structural anchors, and recursive refinement is applied over $n = 2$ iterations to stabilize progressive multimodal fusion. The model is trained using the Adam optimizer with a learning rate of $1 \times 10^{-4}$, batch size of 32, and dropout rate of 0.1. The hybrid gating coefficient $\alpha$ is initialized to 0.5 to ensure balanced contributions from both modalities at the start of training. Notably, this configuration exhibits consistent behavior across datasets. When applied to the MATWI dataset with identical hyperparameters, the model achieves comparable accuracy with only minor performance variations. This suggests that the selected hyperparameters reflect general architectural trade-offs rather than being overfitted to a specific dataset.

\subsection{Experimental Results and Discussion}
\subsubsection{Comparative Experiments}
Tables~\ref{tab:main_results} report the predictive performance on the MATWI and Mudestreda datasets across multiple forecasting horizons.
\textsc{OmniFuser} achieves the best or second-best results in both MSE and MAE across nearly all horizons, demonstrating superior stability in both short- and long-term prediction settings.
On MATWI, it surpasses most baselines at medium-to-long horizons (48-96), reducing the average MSE by over 10\% compared with the strongest competing models. On Mudestreda, where multimodal dependencies are more complex, \textsc{OmniFuser} shows the largest relative improvement at 48-96 steps, confirming its robustness in modeling slow temporal drift and asynchronous visual-temporal interactions. Paired $t$-tests against the top-performing baselines at each horizon yield statistically significant gains ($p<0.01$). These results highlight \textsc{OmniFuser}’s capability to maintain high predictive fidelity over extended horizons while preserving generalization across different industrial datasets.

\begin{figure*}[!t]
\centering
\begin{minipage}{0.49\textwidth}
    \centering
    \includegraphics[width=\linewidth]{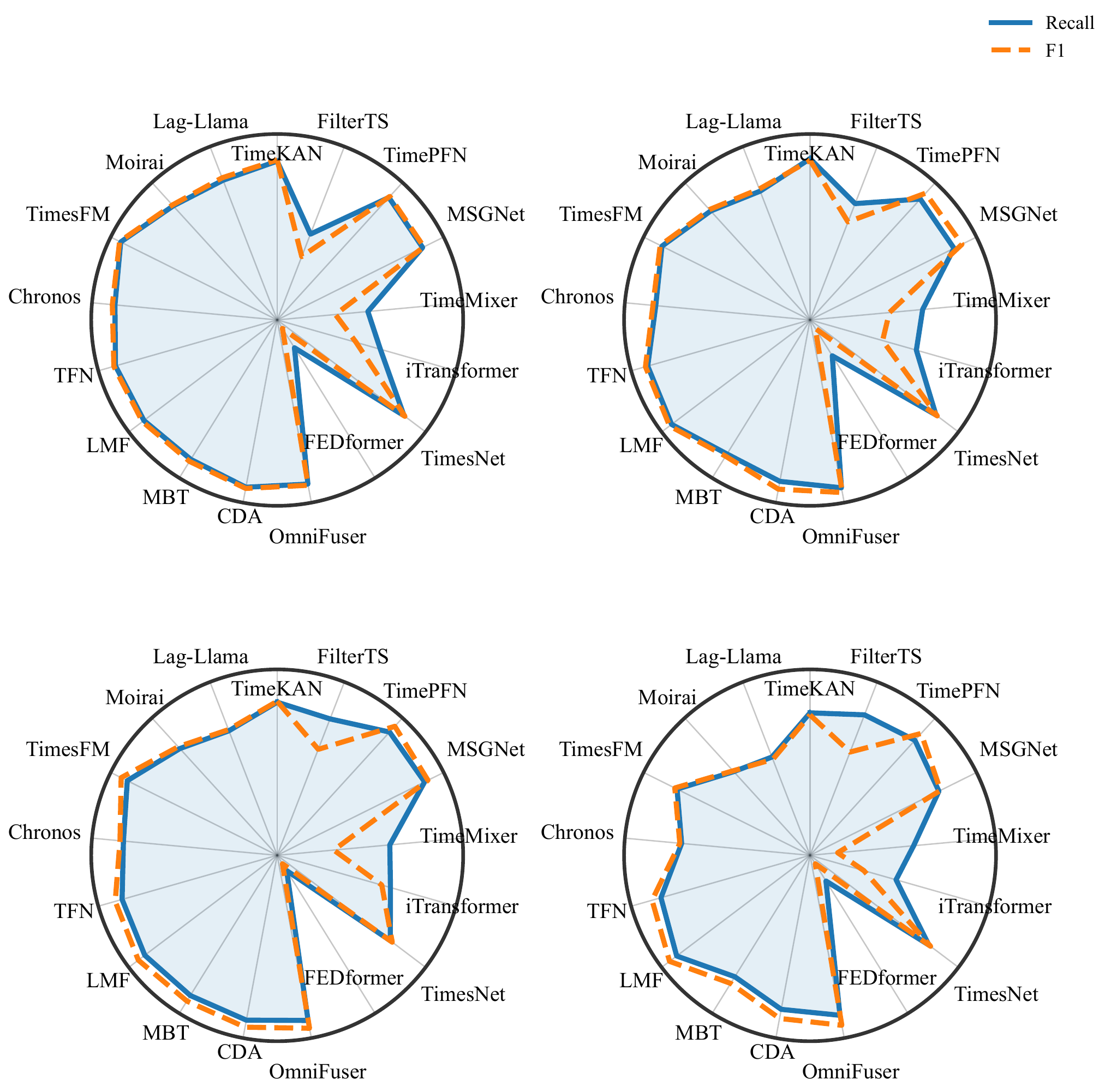}
    \caption{Comparison of all models at horizons 12-48.}
    \label{rader_grid1}
\end{minipage}
\hfill
\begin{minipage}{0.49\textwidth}
    \centering
    \includegraphics[width=\linewidth]{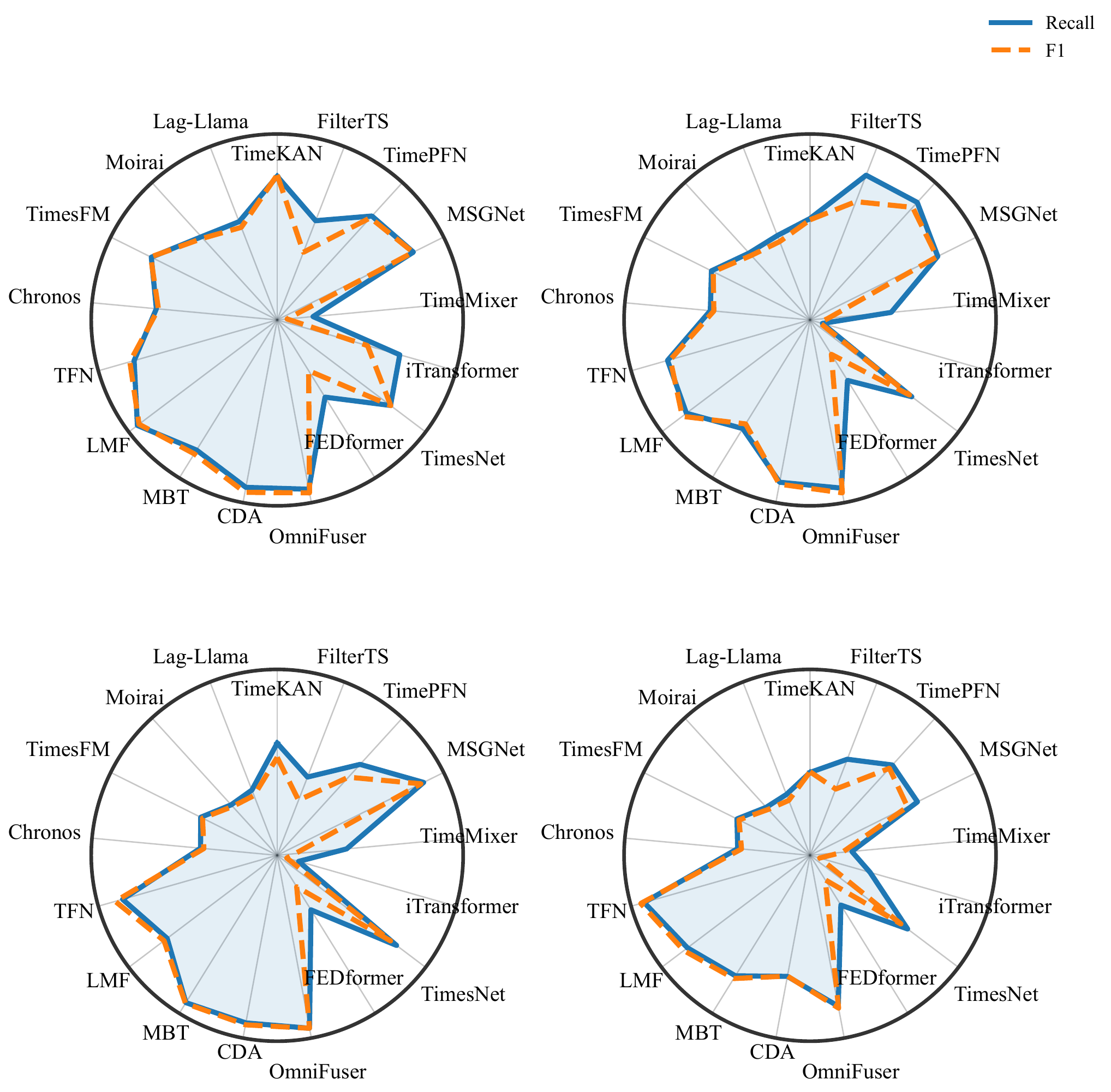}
    \caption{Comparison of all models at horizons 60-96.}
    \label{rader_grid2}
\end{minipage}
\end{figure*}

For classification tasks, \textsc{OmniFuser} consistently achieves top-tier performance across all metrics, including Accuracy, Precision, Recall, and F1, as shown in Tables~\ref{tab:class2} and Figs.~\ref{rader_grid1}-\ref{rader_grid2}.
It remains highly competitive against both time-series models and multimodal baselines, showing stable advantages especially at longer prediction horizons.
On the Mudestreda dataset, \textsc{OmniFuser} attains almost the highest average F1 and Recall among all compared models, reflecting its robustness under multimodal temporal drift and sensor noise.
These findings further validate that \textsc{OmniFuser} generalizes well across both regression and classification objectives, maintaining balanced predictive accuracy and interpretability under diverse multimodal conditions.

\begin{figure}[!ht]
\centering\includegraphics[width=1\columnwidth]{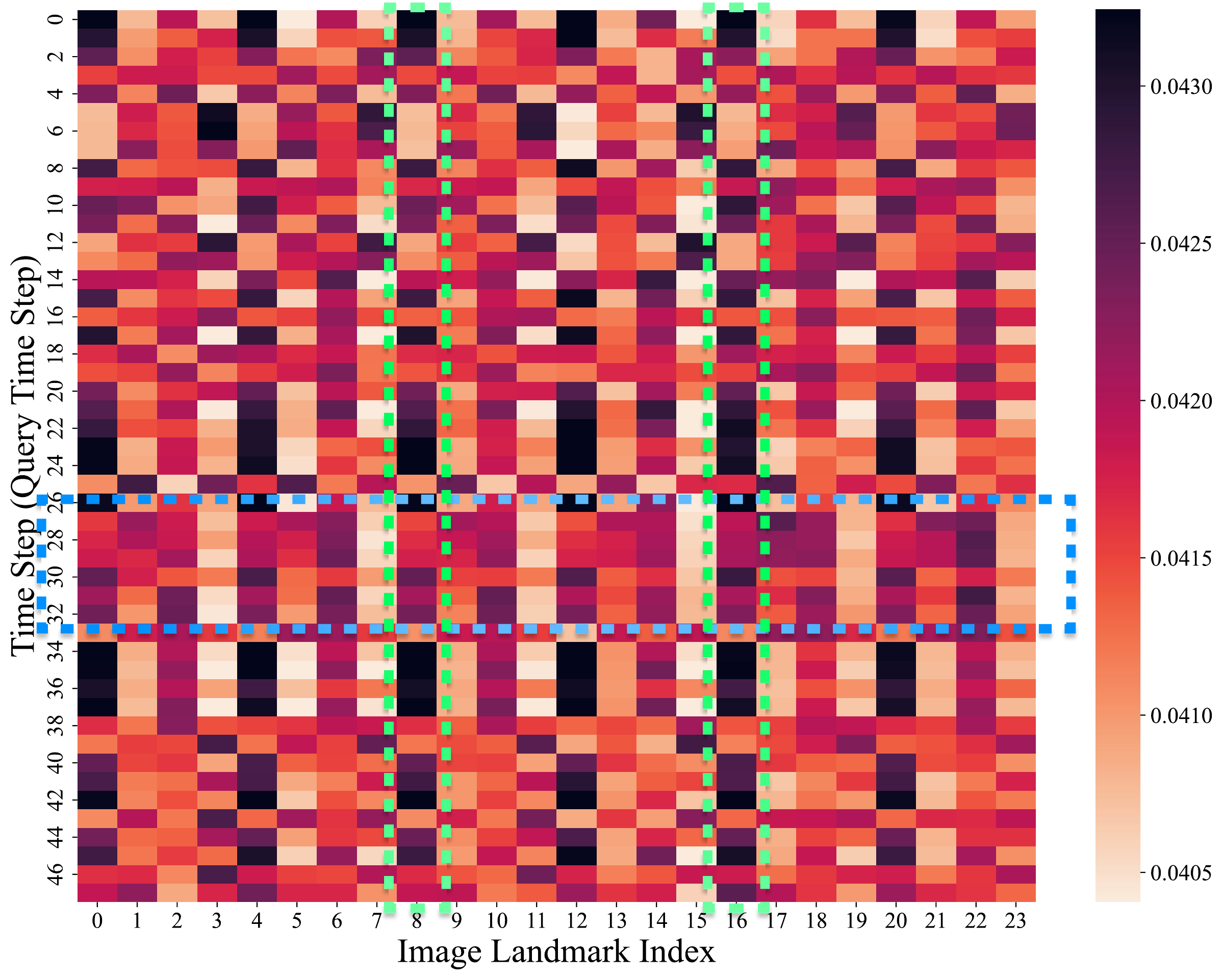}  
\caption{Cross-modal attention dynamics between temporal steps and selected image landmarks.} 
\label{attention}
\end{figure}

Overall, \textsc{OmniFuser} demonstrates consistent and resilient performance across regression and classification tasks under diverse horizons and datasets, underscoring the effectiveness of the proposed contamination-free, low-complexity fusion framework for reliable industrial prediction and decision support. Furthermore, Fig.~\ref{attention} presents the average cross-attention heatmap obtained from ten representative samples, illustrating how the model allocates attention between temporal queries and image landmarks.
Two salient evolution patterns can be observed along the semantic (vertical) and temporal (horizontal) dimensions.
\begin{enumerate}
    \item \textit{Semantic evolution:} distinct vertical stripes indicate that the model exhibits strong selectivity toward visual landmarks. Certain landmarks (e.g., $k = 0,4,8,12,16,20$) consistently receive lower attention across all time steps, suggesting that the model has learned to suppress irrelevant or redundant visual cues.
    \item \textit{Temporal evolution:} alternating horizontal bands show that the fusion strategy evolves over time rather than remaining static. The model adapts its attention allocation across different prediction stages, emphasizing different landmark subsets for short-term versus long-term forecasts.
\end{enumerate}
This joint evolution demonstrates that \textsc{OmniFuser} learns a dynamic, context-aware integration mechanism that captures evolving multimodal dependencies. 

\begin{figure}[!ht]
\centering\includegraphics[width=0.95\columnwidth]{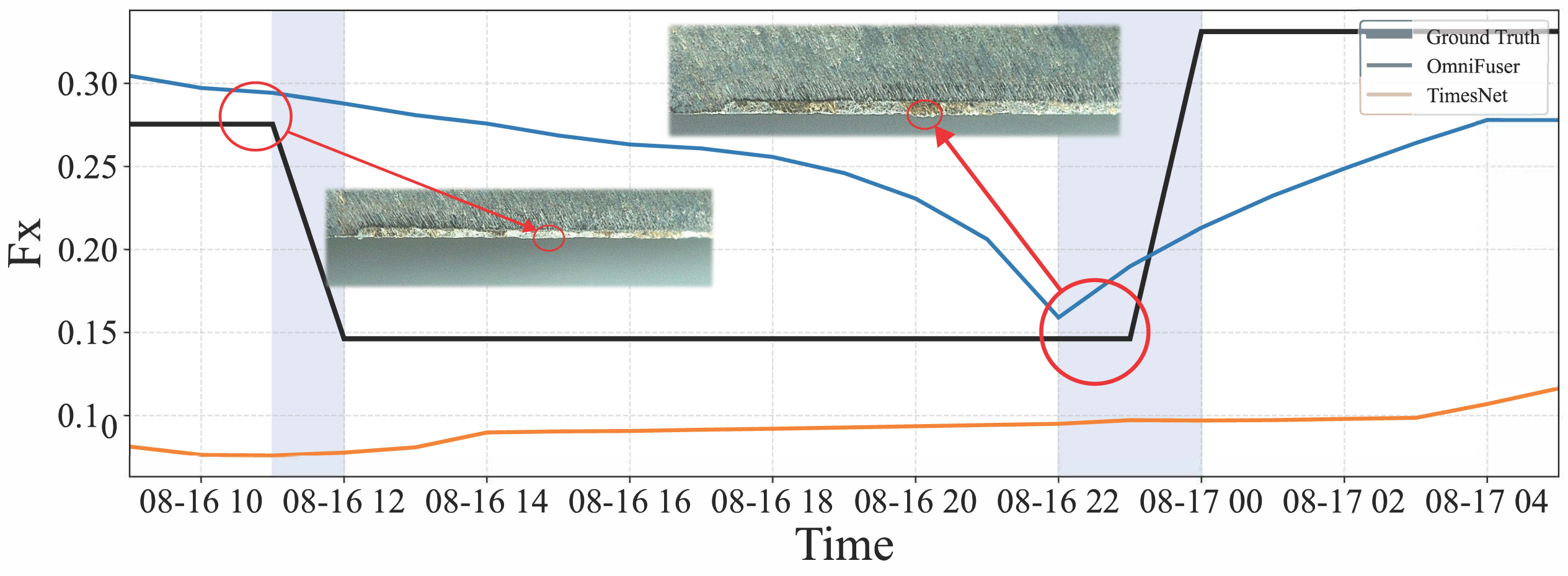}  
\caption{Visual Cues Correct Time-Series Drift in Fx.} 
\label{fig}
\end{figure}

Finally, as shown in Fig.~\ref{fig}, a representative case from the Mudestreda dataset illustrates the evolution of the tangential cutting force $F_x$ over the period from \textit{2018-08-16 12:00} to \textit{2018-08-17 12:00}. 
Around \textit{08-16 22:00}, a sudden regime shift occurs as the ground-truth curve rises sharply after a long steady stage. 
The unimodal time-series model fails to capture this abrupt change and maintains an almost flat prediction, 
whereas \textsc{OmniFuser} promptly turns upward and closely follows the true trajectory. 
This correction arises from its visual branch, which detects wear-induced surface defects in the synchronized tool images (highlighted by red circles) 
and injects these cues into the temporal predictor through cross-modal fusion. 
These visual signals effectively compensate for temporal drift under non-stationary operating conditions. 

\begin{table*}[t!]
\centering
\small
\setlength{\tabcolsep}{2.6pt}
\renewcommand{\arraystretch}{0.6}
\caption{Prediction results under varying horizons. We highlight the best and the second-best results in \textbf{bold} and \underline{underline}.}
\begin{tabular}{c@{\hskip 10pt}c
@{\hskip 6pt}|c@{\hskip 2.5pt}c
@{\hskip 6pt}|c@{\hskip 2.5pt}c
@{\hskip 6pt}|c@{\hskip 2.5pt}c
@{\hskip 6pt}|c@{\hskip 2.5pt}c
@{\hskip 6pt}|c@{\hskip 2.5pt}c
@{\hskip 6pt}|c@{\hskip 2.5pt}c
@{\hskip 6pt}|c@{\hskip 2.5pt}c
@{\hskip 6pt}|c@{\hskip 2.5pt}c
@{\hskip 6pt}|c@{\hskip 2.5pt}c}
\toprule
\multicolumn{2}{c|}{Model} & \multicolumn{2}{c}{TimeKAN} & \multicolumn{2}{c}{FilterTS} & \multicolumn{2}{c}{TimePFN} 
& \multicolumn{2}{c}{MSGNet} & \multicolumn{2}{c}{TimeMixer} &  \multicolumn{2}{c}{iTransformer} & \multicolumn{2}{c}{TimesNet} & \multicolumn{2}{c}{FEDformer} & \multicolumn{2}{c}{OmniFuser} \\
\cmidrule(lr){1-2} \cmidrule(lr){3-4} \cmidrule(lr){5-6} \cmidrule(lr){7-8} \cmidrule(lr){9-10} \cmidrule(lr){11-12} \cmidrule(lr){13-14} \cmidrule(lr){15-16} \cmidrule(lr){17-18}  \cmidrule(lr){19-20} 
\multicolumn{2}{c|}{Metric} & MSE & MAE & MSE & MAE & MSE & MAE & MSE & MAE & MSE & MAE & MSE & MAE & MSE & MAE & MSE & MAE & MSE & MAE\\
\midrule
\multirow{10}{*}{\rotatebox{90}{MATWI}}
& 12  & 0.051 & \textbf{0.043} & 0.059 & 0.060 & 0.058 & 0.097 & 0.051 & \underline{0.044} & 0.126 & 0.098 & 0.182 & 0.135 & 0.109 & 0.183 & \textbf{0.049} & 0.357 & \underline{0.050} & 0.050 \\
& 24 & 0.067 & 0.052 & 0.270 & 0.172 & 0.076 & 0.109 & 0.070 & 0.057 & \underline{0.059} & \textbf{0.048} & 0.173 & 0.130 & 0.221 & 0.158 & \underline{0.059} & 0.237 & \textbf{0.051} & \underline{0.051} \\
& 36 & 0.089 & 0.068 & 0.290 & 0.131 & 0.093 & 0.121 & 0.085 & \underline{0.066} & \underline{0.073} & \textbf{0.055} & 0.169 & 0.127 & 0.202 & 0.148 & 0.076 & 0.269 & \textbf{0.070} & \textbf{0.055} \\
& 48 & 0.104 & \underline{0.078} & 0.198 & 0.222 & 0.112 & 0.135 & \underline{0.102} & \textbf{0.076} & 0.121 & 0.104 & 0.113 & 0.137 & 0.218 & 0.157 & \textbf{0.090} & 0.178 & 0.105 & 0.082 \\
& 60 & 0.124 & \underline{0.090} & 0.718 & 0.367 & 0.130 & 0.115 & 0.123 & \underline{0.090} & 0.131 & 0.101 & \underline{0.119} & \underline{0.086} & 0.153 & \textbf{0.076} & \textbf{0.108} & 0.198 & \textbf{0.108} & \textbf{0.076} \\
& 72 & 0.140 & \underline{0.100} & 0.326 & 0.221 & 0.148 & 0.160 & 0.139 & \underline{0.100} & 0.142 & 0.111 & 0.262 & 0.184 & 0.288 & 0.198 & \underline{0.126} & 0.263 & \textbf{0.125} & \textbf{0.090} \\
& 84 & 0.159 & 0.113 & 0.206 & 0.140 & 0.166 & 0.173 & 0.144 & \textbf{0.099} & 0.193 & 0.145 & \textbf{0.135} & \underline{0.106} & 0.241 & 0.171 & \underline{0.141} & 0.160 & \underline{0.141} & \textbf{0.099} \\
& 96 & 0.179 & 0.126 & 2.259 & 0.680 & 0.183 & 0.186 & 0.164 & \underline{0.113} & 0.181 & 0.132 & \textbf{0.158} & \textbf{0.110} & 0.248 & 0.175 & 0.169 & 0.181 & \underline{0.162} & \underline{0.114} \\
\midrule
\multirow{10}{*}{\rotatebox{90}{Mudestreda}} 
& 12  & 0.681 & 0.598 & 0.701 & 0.569 & \underline{0.557} & \underline{0.524} & 0.669 & 0.592 & 0.679 & 0.595 & 0.592 & 0.536 & 0.672 & 0.595 & 0.694 & 0.622 & \textbf{0.467} & \textbf{0.475} \\
& 24 & 0.752 & 0.635 & 0.849 & 0.646 & 0.696 & 0.603 & 0.746 & 0.627 & 0.749 & 0.627 & \underline{0.677} & \underline{0.585} & 0.747 & 0.633 & 0.745 & 0.632 & \textbf{0.514} & \textbf{0.509} \\
& 36 & 0.833 & 0.672 & 0.948 & 0.692 & 0.792 & 0.643 & 0.824 & 0.665 & 0.835 & 0.671 & \underline{0.774} & \underline{0.635} & 0.822 & 0.667 & 0.776 & 0.652 & \textbf{0.586} & \textbf{0.548} \\
& 48 & 0.919 & 0.703 & 1.046 & 0.731 & 0.890 & 0.683 & 0.912 & 0.697 & 1.082 & 0.828 & 0.859 & \underline{0.669} & 0.904 & 0.698 & \underline{0.856} & 0.676 & \textbf{0.686} & \textbf{0.590} \\
& 60 & 1.007 & 0.735 & 1.151 & 0.768 & 0.981 & 0.715 & 1.006 & 0.724 & 1.046 & 0.736 & \underline{0.962} & \underline{0.706} & 0.991 & 0.728 & 1.095 & 0.758 & \textbf{0.772} & \textbf{0.630} \\
& 72 & 1.086 & 0.762 & 1.217 & 0.791 & 1.069 & 0.749 & 1.091 & 0.755 & 1.086 & 0.762 & \underline{1.040} & \underline{0.733} & 1.069 & 0.754 & 1.188 & 0.791 & \textbf{0.830} & \textbf{0.654} \\
& 84 & 1.159 & 0.788 & 1.292 & 0.818 & 1.145 & 0.774 & 1.160 & 0.781 & 1.175 & 0.789 & \underline{1.121} & \underline{0.764} & 1.136 & 0.780 & 1.266 & 0.819 & \textbf{0.917} & \textbf{0.686} \\
& 96 & 1.228 & 0.814 & 1.372 & 0.847 & 1.213 & 0.801 & 1.231 & 0.808 & 1.239 & 0.824 & \underline{1.195} & \underline{0.793} & 1.206 & 0.807 & 1.331 & 0.844 & \textbf{0.902} & \textbf{0.693} \\
\midrule
\multicolumn{2}{c|}{Model} & \multicolumn{2}{c}{CDA} & \multicolumn{2}{c}{MBT} & \multicolumn{2}{c}{LMF} 
& \multicolumn{2}{c}{TFN} & \multicolumn{2}{c}{Chronos} &  \multicolumn{2}{c}{TimesFM} & \multicolumn{2}{c}{Moirai} & \multicolumn{2}{c}{Lag-Llama} & \multicolumn{2}{c}{OmniFuser} \\
\cmidrule(lr){1-2} \cmidrule(lr){3-4} \cmidrule(lr){5-6} \cmidrule(lr){7-8} \cmidrule(lr){9-10} \cmidrule(lr){11-12} \cmidrule(lr){13-14} \cmidrule(lr){15-16} \cmidrule(lr){17-18}  \cmidrule(lr){19-20} 
\multicolumn{2}{c|}{Metric} & MSE & MAE & MSE & MAE & MSE & MAE & MSE & MAE & MSE & MAE & MSE & MAE & MSE & MAE & MSE & MAE & MSE & MAE\\
\midrule
\multirow{10}{*}{\rotatebox{90}{MATWI}}
& 12  & 0.083 & 0.073 & 0.069 & \underline{0.059} & 0.085 & 0.073 & \underline{0.055} & \textbf{0.045} & 1.251 & 0.813 & 1.002 & 0.790 & 3.514 & 1.069 & 1.213 & 0.807 & \textbf{0.050} & \underline{0.050} \\
& 24 & 0.092 & 0.072 & 0.071 & 0.065 & 0.115 & 0.094 & \underline{0.058} & \textbf{0.049} & 1.297 & 0.817 & 1.000 & 0.788 & 2.588 & 1.067 & 1.166 & 0.797 & \textbf{0.051} & \underline{0.051} \\
& 36 & 0.085 & 0.068 & 0.166 & 0.131 & 0.145 & 0.110 & \underline{0.071} & \underline{0.058} & 1.321 & 0.819 & 0.999 & 0.787 & 4.392 & 1.113 & 1.130 & 0.788 & \textbf{0.070} & \textbf{0.055} \\
& 48 & \textbf{0.089} & \underline{0.089} & 0.129 & \underline{0.098} & 0.119 & 0.154 & 0.222 & 0.157 & 1.364 & 0.825 & 1.001 & 0.801 & 2.608 & 1.078 & 1.091 & 0.779 & \underline{0.105} & \textbf{0.082} \\
& 60 & \underline{0.141} & \underline{0.108} & 0.147 & 0.114 & 0.410* & 0.269 & 0.352 & 0.223 & 1.409 & 0.831 & 1.000 & 0.784 & 2.717 & 1.097 & 1.073 & 0.779 & \textbf{0.108} & \textbf{0.076} \\
& 72 & 0.158 & 0.139 & 0.173 & \underline{0.127} & 0.164 & 0.150 & \underline{0.150} & \underline{0.124} & 1.433 & 0.834 & 1.000 & 0.784 & 4.115 & 1.116 & 1.047 & 0.772 & \textbf{0.125} & \textbf{0.090} \\
& 84 & 0.292 & 0.195 & \underline{0.144} & 0.169 & 0.152 & \underline{0.118} & 0.235 & 0.171 & 1.449 & 0.837 & 0.999 & 0.786 & 3.168 & 1.124 & 1.036 & 0.772 & \textbf{0.142} & \textbf{0.099} \\
& 96 & 0.243 & 0.191 & 0.282 & 0.215 & 0.494 & 0.283 & \underline{0.165} & \underline{0.118} & 1.469 & 0.841 & 1.000 & 0.784 & 17.402 & 1.177 & 1.015 & 0.769 & \textbf{0.162} & \textbf{0.114} \\
\midrule
\multirow{10}{*}{\rotatebox{90}{Mudestreda}}
& 12  & 0.650 & 0.583 & 0.498 & 0.483 & \underline{0.472} & \underline{0.468} & 0.506 & 0.480 & 0.516 & \textbf{0.439} & 0.656 & 0.489 & 1.166 & 0.735 & 0.629 & 0.539 & \textbf{0.467} & 0.475 \\
& 24 & 0.724 & 0.617 & 0.560 & 0.527 & 0.572 & 0.528 & \underline{0.535} & \underline{0.513} & 0.579 & \textbf{0.496} & 0.708 & 0.549 &  1.665 & 0.825 & 0.692 & 0.588 & \textbf{0.514} & \underline{0.509} \\
& 36 & 0.811 & 0.660 & \underline{0.600} & \underline{0.551} & 0.618 & 0.560 & 0.622 & 0.567 & 0.681 & 0.557 & 0.803 & 0.607 & 1.390 & 0.821 & 0.769 & 0.625 & \textbf{0.586} & \textbf{0.548} \\
& 48 & 0.904 & 0.690 & 0.709 & 0.596 & \textbf{0.657} & \textbf{0.580} & \underline{0.662} & \underline{0.584} & 0.754 & 0.594 & 0.876 & 0.646 & 1.554 & 0.863 & 0.861 & 0.656 & 0.686 & 0.590 \\
& 60 & 0.993 & 0.726 & 0.883 & 0.662 & \underline{0.754} & 0.622 & \textbf{0.728} & \underline{0.611} & 0.863 & 0.640 & 0.966 & 0.686 & 1.637 & 0.886 & 0.825 & \textbf{0.607} & \underline{0.772} & 0.630 \\
& 72 & 0.877 & \textbf{0.595} & \textbf{0.795} & \underline{0.641} & 0.850 & 0.664 & 0.835 & \underline{0.654} & 0.943 & 0.671 & 1.032 & 0.716 & 1.657 & 0.909 & 1.056 & 0.721 & \underline{0.830} & \underline{0.654} \\
& 84 & 1.090 & 0.766 & 0.883 & \underline{0.677} & \underline{0.878} & 0.679 & \textbf{0.855} & \textbf{0.667} & 1.028 & 0.705 & 1.108 & 0.746 & 1.586 & 0.914 & 1.151 & 0.750 & 0.917 & 0.686 \\
& 96 & 1.127 & 0.782 & 0.950 & 0.703 & \underline{0.938} & \underline{0.701} & 0.942 & 0.893 & 1.100 & 0.735 & 1.179 & 0.774 & 1.806 & 0.955 & 1.227 & 0.777 & \textbf{0.902} & \textbf{0.693} \\
\bottomrule
\end{tabular}
\label{tab:main_results}
\end{table*}

\begin{table*}[t!]
\centering
\small
\setlength{\tabcolsep}{2.6pt}
\renewcommand{\arraystretch}{0.6}
\caption{Classification results on two real-world datasets under varying prediction horizons.} 
\begin{tabular}{c@{\hskip 10pt}c
@{\hskip 6pt}|c@{\hskip 2.5pt}c
@{\hskip 6pt}|c@{\hskip 2.5pt}c
@{\hskip 6pt}|c@{\hskip 2.5pt}c
@{\hskip 6pt}|c@{\hskip 2.5pt}c
@{\hskip 6pt}|c@{\hskip 2.5pt}c
@{\hskip 6pt}|c@{\hskip 2.5pt}c
@{\hskip 6pt}|c@{\hskip 2.5pt}c
@{\hskip 6pt}|c@{\hskip 2.5pt}c
@{\hskip 6pt}|c@{\hskip 2.5pt}c}
\toprule
\multicolumn{2}{c|}{Model} & \multicolumn{2}{c}{TimeKAN} & \multicolumn{2}{c}{FilterTS} & \multicolumn{2}{c}{TimePFN} 
& \multicolumn{2}{c}{MSGNet} & \multicolumn{2}{c}{TimeMixer} &  \multicolumn{2}{c}{iTransformer} & \multicolumn{2}{c}{TimesNet} & \multicolumn{2}{c}{FEDformer} & \multicolumn{2}{c}{OmniFuser} \\
\cmidrule(lr){1-2} \cmidrule(lr){3-4} \cmidrule(lr){5-6} \cmidrule(lr){7-8} \cmidrule(lr){9-10} \cmidrule(lr){11-12} \cmidrule(lr){13-14} \cmidrule(lr){15-16} \cmidrule(lr){17-18}  \cmidrule(lr){19-20} 
\multicolumn{2}{c|}{Metric} & Acc & Pre & Acc & Pre & Acc & Pre & Acc & Pre & Acc & Pre & Acc & Pre & Acc & Pre & Acc & Pre & Acc & Pre\\
\midrule
\multirow{10}{*}{\rotatebox{90}{Mudestreda}} 
& 12  & 0.860 & 0.869 & 0.515 & 0.505 & \textbf{0.897} & \textbf{0.898} & 0.879 & 0.884 & 0.567 & 0.352 & 0.564 & 0.579 & 0.860 & 0.869 & 0.276 & 0.186 & \underline{0.886} & \underline{0.890} \\
& 24 & 0.843 & 0.824 & 0.632 & 0.654 & \underline{0.917} & \underline{0.912} & 0.878 & 0.885 & 0.582 & 0.421 & 0.532 & 0.651 & 0.823 & 0.838 & 0.222 & 0.187 & \textbf{0.928} & \textbf{0.922} \\
& 36 & 0.768 & 0.733 &  0.630 & 0.714 & \underline{0.832} & \underline{0.825} & 0.814 & 0.802 & 0.489 & 0.351 & 0.726 & 0.772 & 0.719 & 0.710 & 0.265 & 0.277 & \textbf{0.838} & \textbf{0.905} \\
& 48 & 0.755 & 0.684 & 0.623 & 0.677 & \underline{0.851} & \underline{0.806} & 0.767 & 0.710 & 0.441 & 0.294 & 0.450 & 0.553 & 0.783 & 0.741 & 0.346 & 0.353 & \textbf{0.866} & \textbf{0.913} \\
& 60 & 0.692 & 0.676 & 0.433 & 0.489 & 0.630 & 0.616 & \underline{0.727} & \underline{0.684} & 0.347 & 0.290 & 0.488 & 0.547 & 0.673 & 0.637 & 0.452 & 0.447 & \textbf{0.820} & \textbf{0.802} \\
& 72 & 0.587 & 0.529 & 0.616 & 0.635 & \underline{0.712} & \underline{0.678} & 0.696 & \underline{0.687} & 0.325 & 0.225 & 0.260 & 0.258 & 0.653 & 0.622 & 0.382 & 0.413 & \textbf{0.863} & \textbf{0.862} \\
& 84 & 0.565 &  0.603 & 0.422 & 0.431 & 0.570 & 0.570 & \underline{0.760} & \underline{0.780} & 0.335 & 0.250 & 0.287 & 0.243 & 0.689 & 0.708 & 0.391 & 0.400 & \textbf{0.869} & \textbf{0.853} \\
& 96 & 0.564 & 0.499 & 0.469 & 0.521 & \underline{0.648} & \underline{0.676} & 0.622 & 0.658 & 0.458 & 0.321 & 0.309 & 0.579 & 0.627 & \underline{0.639} & 0.358 & 0.385 & \textbf{0.863} & \textbf{0.842} \\
\midrule
\multicolumn{2}{c|}{Model} & \multicolumn{2}{c}{CDA} & \multicolumn{2}{c}{MBT} & \multicolumn{2}{c}{LMF} 
& \multicolumn{2}{c}{TFN} & \multicolumn{2}{c}{Chronos} &  \multicolumn{2}{c}{TimesFM} & \multicolumn{2}{c}{Moirai} & \multicolumn{2}{c}{Lag-Llama} & \multicolumn{2}{c}{OmniFuser} \\
\cmidrule(lr){1-2} \cmidrule(lr){3-4} \cmidrule(lr){5-6} \cmidrule(lr){7-8} \cmidrule(lr){9-10} \cmidrule(lr){11-12} \cmidrule(lr){13-14} \cmidrule(lr){15-16} \cmidrule(lr){17-18}  \cmidrule(lr){19-20} 
\multicolumn{2}{c|}{Metric} & Acc & Pre & Acc & Pre & Acc & Pre & Acc & Pre & Acc & Pre & Acc & Pre & Acc & Pre & Acc & Pre & Acc & Pre\\
\midrule
\multirow{10}{*}{\rotatebox{90}{Mudestreda}}
& 12  & \underline{0.918} & \textbf{0.916} & 0.902 & 0.902 & 0.905 & 0.911 & 0.910 & 0.911 & 0.895 & 0.903 & \textbf{0.934} & \underline{0.933} & 0.853 & 0.849 & 0.845 & 0.851 & 0.886 & 0.889 \\
& 24 & 0.898 & 0.890 & 0.833 & 0.873 & \underline{0.911} & \underline{0.918} & 0.888 & 0.904 & 0.844 & 0.845 & 0.879 & 0.874 & 0.808 & 0.798 & 0.782 & 0.778 & \textbf{0.928} & \textbf{0.922} \\
& 36 & \underline{0.830} & \underline{0.866} & 0.825 & 0.855 & 0.824 & \underline{0.874} & 0.806 & 0.853 & 0.781 & 0.769 & 0.824 & 0.808 & 0.745 & 0.722 & 0.724 & 0.698 & \textbf{0.838} & \textbf{0.905} \\
& 48 & 0.813 & 0.822 & 0.811 & 0.846 & \underline{0.849} & \underline{0.851} & 0.842 & \underline{0.896} & 0.723 & 0.686 & 0.766 & 0.733 & 0.684 & 0.628 & 0.668 & 0.607 & \textbf{0.866} & \textbf{0.913} \\
& 60 & 0.808 & \textbf{0.870} & 0.768 & 0.835 & \underline{0.818} & 0.806 & 0.755 & \underline{0.851} & 0.663 & 0.588 & 0.708 & 0.648 & 0.639 & 0.547 & 0.621 & 0.518 & \textbf{0.820} & 0.802 \\
& 72 & 0.807 & 0.741 & 0.771 & 0.660 & 0.828 & \underline{0.880} & \underline{0.835} & \textbf{0.901} & 0.622 & 0.523 & 0.653 & 0.558 & 0.591 & 0.487 & 0.588 & 0.480 & \textbf{0.863} & 0.862 \\
& 84 & 0.830 & 0.839 & 0.828 & 0.843 & 0.770 & 0.786 & \textbf{0.874} & \textbf{0.914} & 0.582 & 0.485 & 0.614 & 0.518 & 0.547 & 0.449 & 0.553 & 0.448 & \underline{0.869} & \underline{0.853} \\
& 96 & 0.755 & 0.707 & 0.712 & 0.746 & 0.819 & 0.769 & \underline{0.842} & \underline{0.799} & 0.544 & 0.450 & 0.574 & 0.481 & 0.520 & 0.422 & 0.517 & 0.418 & \textbf{0.863} & \textbf{0.842} \\
\bottomrule
\end{tabular}
\label{tab:class2}
\end{table*}

\subsubsection{Abalation Study}
\begin{figure}[!ht]
\centering
\includegraphics[width=0.98\columnwidth]{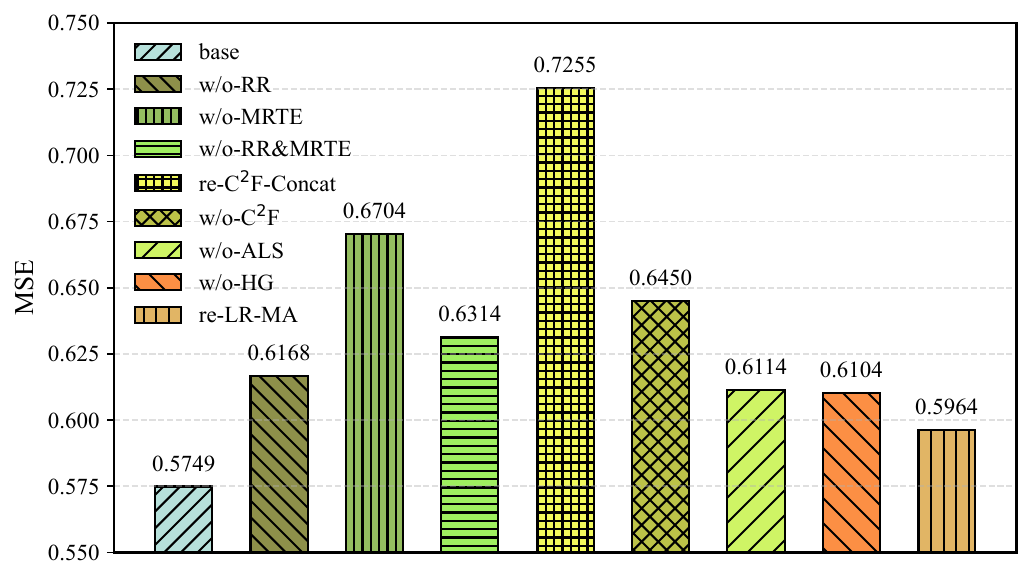} 
\caption{Ablation study on two datasets (average results are reported).}
\label{abalation_mse}
\end{figure}

To assess the contribution of each core component in \textsc{OmniFuser}, we conduct detailed ablation studies. Fig.~\ref{abalation_mse} reports the MSE of different variants. Removing Recursive Refinement (w/o-RR) significantly increases error, confirming its role in retaining residual information and stabilizing fusion dynamics. Excluding multi-resolution decomposition in MRTE (w/o-MRTE) leads to noticeable degradation, as the model loses the ability to capture long- and short-term temporal dependencies. Removing both (w/o-RR\&MRTE) yields a lower error than removing MRTE alone. This is because RR relies on MRTE’s decomposition. Without MRTE, the recursive updates mainly recycle noisy single-scale features and amplify errors. Disabling both simultaneously avoids this mismatch, leading to a slightly better but still inferior result compared to the full model. Replacing the proposed $\text{C}^2\text{F}$ module with a simple feature concatenation (re-$\text{C}^2\text{F}$-Concat) or removing it entirely (w/o-$\text{C}^2\text{F}$) causes substantial accuracy loss, showing the necessity of contamination-free cross-modal interaction. To validate Theorem~2 empirically, we estimate the kernel’s effective rank, confirm a $>0.9$ correlation between ALS and leverage scores, and find that $k=2$ achieves the lowest error in Fig.~\ref{abalation_mse1}, indicating an optimal balance between approximation accuracy and efficiency. Within $\text{C}^2\text{F}$, removing Adaptive Landmark Selection (w/o-ALS) reduces fidelity of cross-modal attention, while discarding the Hybrid Gate (w/o-HG) undermines the balance between global trends and local fluctuations. Finally, substituting decomposition with a moving average (re-LR-MA) weakens the discrimination of tool degradation patterns. Overall, the full model consistently outperforms all variants, validating that each module provides complementary benefits to predictive maintenance performance.

\subsubsection{Computational Cost}
\begin{figure}[!ht]
\centering\includegraphics[width=0.92\columnwidth]{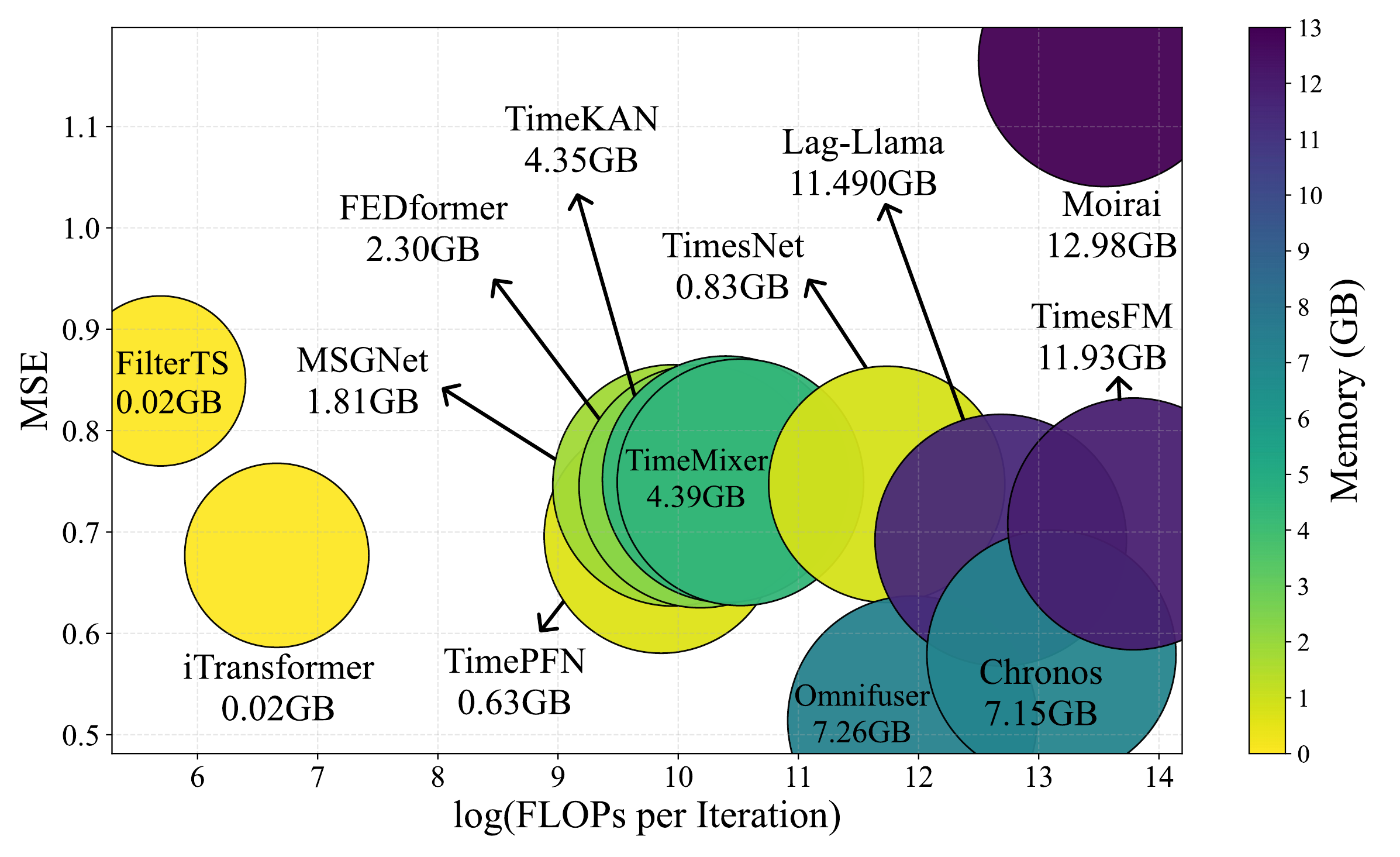} 
\caption{Performance Analysis of OmniFuser.} 
\label{fig:efficiency}
\end{figure}

Fig.~\ref{fig:efficiency} illustrates the trade-off between computational cost, memory footprint, and predictive accuracy of different models on the Mudestreda dataset.
\textsc{OmniFuser} achieves the lowest MSE while maintaining moderate complexity, requiring approximately $10^{11}$~FLOPs per iteration and 7.26~GB of memory.
This positions \textsc{OmniFuser} among the models that achieve a favorable balance between accuracy and computational efficiency.
Although \textsc{OmniFuser} incorporates an additional image modality, its efficient fusion structure and compact visual encoder prevent an exponential increase in computational demand.
Compared with large multimodal or long-sequence transformers, which often exceed 10-12~log(FLOPs) and demand more than 10~GB of memory, \textsc{OmniFuser} reduces the per-iteration cost by more than 40\% while achieving superior accuracy.
These results demonstrate that its framework effectively leverages multimodal information without compromising computational scalability, making \textsc{OmniFuser} well-suited for deployment in most data-intensive industrial environments.

\subsubsection{Discussion}

In the milling scenario, \textsc{OmniFuser} is well-suited for real-world deployment by integrating common force or vibration sensors with a low-cost industrial camera mounted near the tool. 
Modern CNC machines already support such sensor-camera setups \cite{CNC}, and the gradual nature of tool wear ensures that low-frequency imaging remains sufficient. 
This makes the framework practical and effective for capturing multimodal tool conditions without disrupting production. 
In addition, this setting is feasible in practice, as in industrial machining, a tool usually processes a batch of similar workpieces continuously until replacement, making its wear trajectory a naturally continuous process for prediction. The model maintains a moderate power footprint of around 110~W on industrial GPUs such as the NVIDIA RTX~A4000 or L4 \cite{GPU}, making it suitable for continuous on-machine deployment in real-world machining scenarios. This value corresponds to the typical power draw observed when utilizing about 30-40\% of GPU resources for inference, given the model’s computational complexity of roughly $10^{11}$~FLOPs per iteration. 

\textsc{OmniFuser} also exhibits strong robustness under partial modality absence, which is common in industrial environments where sensors may fail or images become occluded by coolant or chips. 
Owing to the contamination-free $\text{C}^2\text{F}$ design, the private-shared decomposition limits cross-modal interference and allows the remaining modality to preserve its discriminative subspace. 
When the visual input is unavailable, the recursive refinement path can still recycle and propagate residual information from the sensor modality, maintaining stable forecasting performance. 
Conversely, when sensor noise corrupts the signal, the visual stream provides slow-varying contextual priors that anchor the degradation trend, leading to graceful performance degradation rather than abrupt collapse. 
In practice, we adopt standard modality dropout during training and masking at inference to ensure this behavior \cite{drop}, making it particularly practical for real-time industrial deployment.

Furthermore, it is broadly applicable beyond the specific case of tool wear monitoring. Its central contribution lies in establishing a principled mechanism for aligning and fusing heterogeneous modalities, which can be encapsulated as a reusable component within intelligent maintenance service frameworks. This property makes the approach suitable for a wide range of predictive maintenance tasks where visual inspections and sensor readings can be jointly analyzed. For example, in bearing monitoring, periodic thermal or microscopic images can be paired with vibration measurements to capture both surface degradation and dynamic response. In turbine systems, endoscopic images obtained at inspection intervals can be fused with continuous acoustic emissions to link internal structural defects with operational signals. In production lines, camera snapshots of material flow can be integrated with load or torque measurements to provide a holistic view of system dynamics within service-oriented predictive maintenance platforms.

\section{Conclusions and Future Work}
\label{CF}
This work proposes \textsc{OmniFuser}, an omnidirectional multimodal fusion framework tailored for service-oriented predictive maintenance in industrial scenarios. By jointly leveraging high-resolution tool images and sensor signals, the model captures complementary spatial and temporal degradation patterns. The proposed Contamination-free Cross-modal Fusion ($\text{C}^2\text{F}$) integrates modality-specific and shared representations, while recursive refinement stabilizes fusion and mitigates information loss. Extensive experiments on real-world datasets demonstrate that \textsc{OmniFuser} consistently outperforms state-of-the-art baselines in both tool wear classification and multi-step force forecasting.

Beyond milling tools, the framework can serve as a reusable component within intelligent maintenance service architectures and be extended to other industrial assets.
Future work will focus on scaling \textsc{OmniFuser} into lightweight variants and enabling service-oriented deployment to ensure real-time, robust, and widely deployable predictive maintenance services.

\bibliography{ref}
\bibliographystyle{IEEEtran}     

\end{document}